\documentclass{article}

\usepackage[final,nonatbib]{neurips_2024}

\usepackage[utf8]{inputenc} 
\usepackage[T1]{fontenc}    
\usepackage{hyperref}       
\usepackage{url}            
\usepackage{booktabs}       
\usepackage{amsfonts}       
\usepackage{nicefrac}       
\usepackage{microtype}      
\usepackage{xcolor}         
\usepackage{multirow}
\usepackage{booktabs}   
\usepackage{bm}
\usepackage{subfigure}
\usepackage{graphicx}
\usepackage{algorithm}
\usepackage{algorithmic}
\usepackage{upgreek}
\usepackage{bm}
\usepackage{amsmath}

\usepackage{subcaption}
\usepackage{natbib}
\setcitestyle{numbers,square}

\title{Reinforcement Learning Policy as Macro Regulator Rather than Macro Placer}

\author{%
  Ke Xue$^{1,2}$, Ruo-Tong Chen$^{1,2}$, Xi Lin$^{1,2}$, Yunqi Shi$^{1,2}$,  \\
  \textbf{Shixiong Kai$^{3}$, Siyuan Xu$^{3}$, Chao Qian}\thanks{Correspondence to Chao Qian \textless\href{mailto:qianc@nju.edu.cn}{qianc@nju.edu.cn}\textgreater}~~$^{1,2}$\\ 
$^1$National Key Laboratory for Novel Software Technology, Nanjing University \\
$^2$School of Artificial Intelligence, Nanjing University  \\
$^3$Huawei Noah’s Ark Lab \\
}

\begin{document}

\maketitle

\begin{abstract}
In modern chip design, placement aims at placing millions of circuit modules, which is an essential step that significantly influences power, performance, and area (PPA) metrics. Recently, reinforcement learning (RL) has emerged as a promising technique for improving placement quality, especially macro placement. However, current RL-based placement methods suffer from long training times, low generalization ability, and inability to guarantee PPA results. A key issue lies in the problem formulation, i.e., using RL to place from scratch, which results in limits useful information and inaccurate rewards during the training process. In this work, we propose an approach that utilizes RL for the refinement stage, which allows the RL policy to learn how to adjust existing placement layouts, thereby receiving sufficient information for the policy to act and obtain relatively dense and precise rewards. Additionally, we introduce the concept of regularity during training, which is considered an important metric in the chip design industry but is often overlooked in current RL placement methods. We evaluate our approach on the ISPD 2005 and ICCAD 2015 benchmark, comparing the global half-perimeter wirelength and regularity of our proposed method against several competitive approaches. Besides, we test the PPA performance using commercial software, showing that RL as a regulator can achieve significant PPA improvements. Our RL regulator can fine-tune placements from any method and enhance their quality. Our work opens up new possibilities for the application of RL in placement, providing a more effective and efficient approach to optimizing chip design. Our code is available at \url{https://github.com/lamda-bbo/macro-regulator}.
\end{abstract}

\section{Introduction}
In the complex and evolving landscape of modern chip design, placement is a pivotal process that significantly influences the power, performance, and area (PPA) metrics of the final chip~\cite{mac2000industrial,markov2012progress}. 
A modern chip typically comprises thousands of macros (i.e., individual building blocks such as memories) and millions of standard cells (i.e., smaller basic components like logic gates). The macro placement result provides a fundamental solution for the subsequent processes (e.g., standard cells placement and routing), thus playing an important role~\cite{tang2007memetic}. 
For example, macro placement influences the placement of standard cells, and poor macro placement might make it challenging to place these cells optimally, leading to an unsatisfactory chip performance~\cite{vashisht2020placement}.
Moreover, an inappropriate macro placement can result in macro blockage in the core center, which harms the overall chip performance by causing unwanted effects such as routing congestion, inferior wirelength, and timing performance issues~\cite{incre-macro}. 

Due to the lengthy and complex workflow of chip design, designers often rely on proxy metrics that can reflect the final results to guide the optimization process~\citep{caldwell1999wirelength,spindler2007fast,lu2015eplace}. One important proxy metric is half-perimeter wirelength (HPWL), which provides an approximation for the routing wirelength and is widely used to measure the placement quality~\cite{caldwell1999wirelength,kahng2006tale,shahookar1991vlsi}. Traditional macro placement methods can be divided into two categories. Earlier approaches usually solve macro placement by black-box optimization (BBO)~\cite{murata1996vlsi,chang2000b,hong2000corner,wiremask-bbo}. They often suffer from the poor scalability due to the large-scale search space and high complexity of decoding a solution to a placement. 
Another type is analytical method~\cite{chen2008ntuplace3,cheng2018replace,lin2020dreamplace}, which can solve the placement efficiently by approximating HPWL gradients. However, these methods are hard to guarantee the non-overlapping constrain between cells and are easy to be stuck in local optima~\citep{lai2022maskplace,xue2024escaping}.

Reinforcement learning (RL) \citep{sutton2018reinforcement} has recently emerged as a promising technique to enhance the macro placement quality \citep{nature-graph,deeppr,cheng2022the,lai2022maskplace,lai2023chipformer}. RL's ability to learn policies through interaction with a complex environment offers a novel pathway for addressing various challenges of macro placement. However, the application of RL is currently hindered by several limitations, including the long training time, an inability to guarantee PPA improvements, and the lack of generalization across different chip layouts. In this work, we highlight that a major contributing factor to these issues is the problem formulation, i.e., the conventional RL approach of placing macros from scratch often results in limited state information and inaccurate reward signal throughout the learning process.


\begin{figure}[t!]
    \begin{minipage}{0.44\textwidth}
        \centering
        \begin{subfigure}
            \centering
            \includegraphics[width=\linewidth]{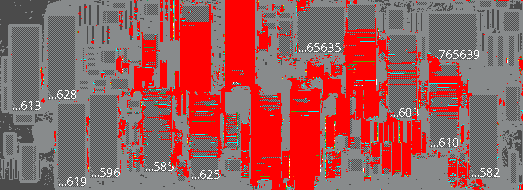}
            \centering
            \text{(a) MaskPlace}
            \vspace{0.65cm}
        \end{subfigure}
        
        \begin{subfigure}
        \centering
        \includegraphics[width=\linewidth]{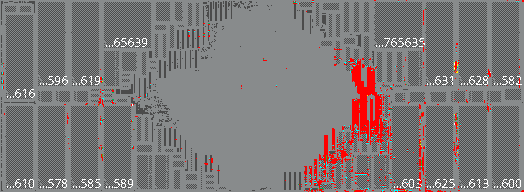} \\
        \centering
        \text{(b) MaskRegulate}
        \end{subfigure}
    \end{minipage}
    \medskip
    \begin{minipage}{0.55\textwidth}
        \centering
        \begin{subfigure}
            \centering
            \includegraphics[width=\linewidth]{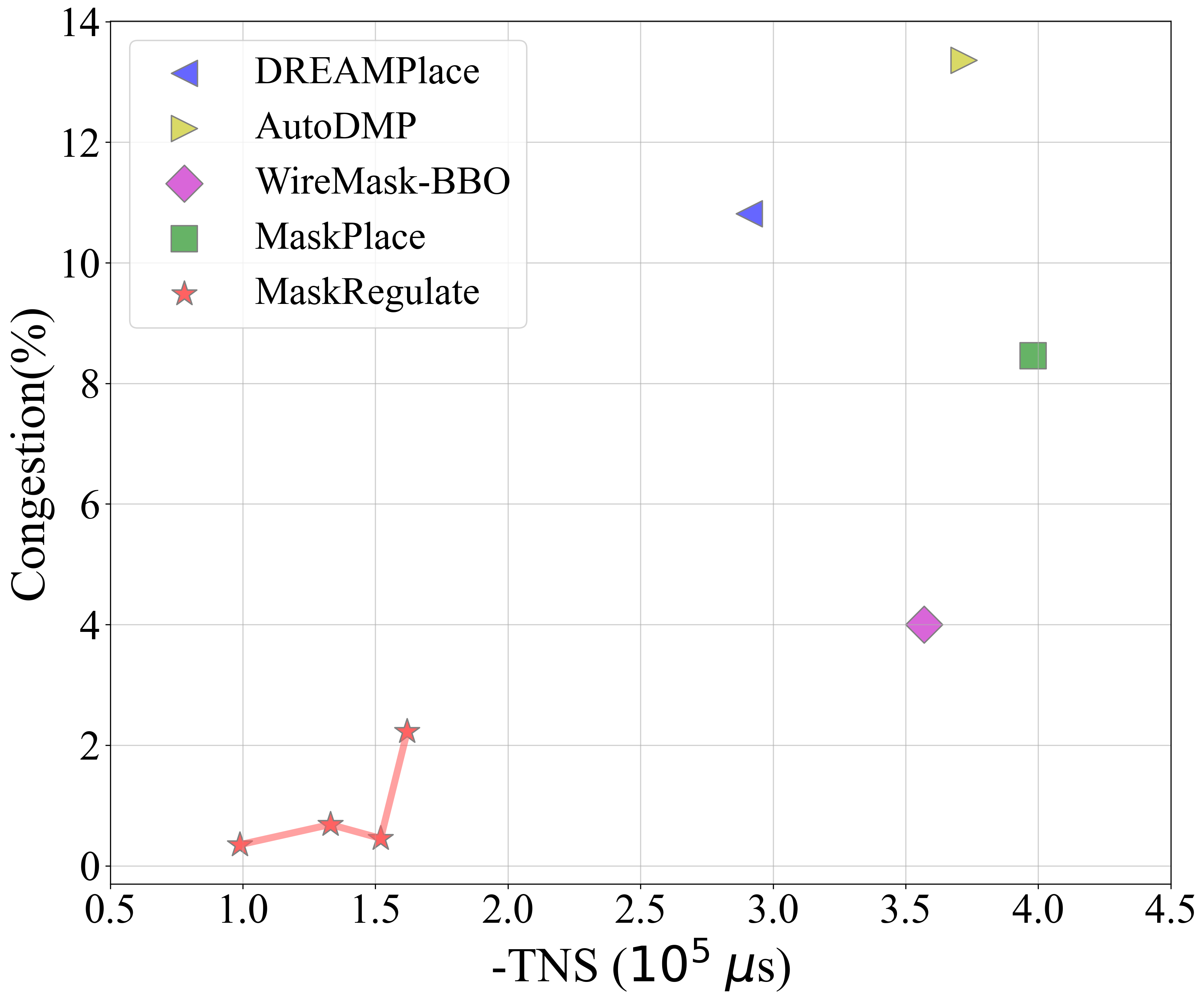}
             \centering
            \text{(c) PPA metrics comparison}
        \end{subfigure}
    \end{minipage}
\caption{Placement layouts and congestions of (a) MaskPlace and (b) MaskRegulate on the superblue1 from ICCAD 2015 benchmark~\citep{iccad15}, where the red points indicate the congestion critical regions. (c): Comparing two crucial PPA metrics, namely Congestion and total negative slack (TNS) between MaskRegulate, DREAMPlace~\citep{lin2020dreamplace}, AutoDMP~\citep{agnesina2023autodmp}, WireMask-BBO~\citep{wiremask-bbo}, and MaskPlace~\citep{lai2022maskplace}, where lower values indicate better performance.}
\label{fig:congestion-fig1}
\end{figure}
To address these challenges, we propose a novel RL approach called MaskRegulate that shifts the focus from initial placement to refining existing placement layouts. The RL policy acts as a regulator rather than a placer, which operates on pre-existing placements, thus allowing for access to comprehensive state information and enabling the acquisition of more precise rewards. This adjustment enhances the efficiency of the learning process and finally improves the final placement results. Furthermore, MaskRegulate introduces the concept of regularity~\citep{incre-macro} as a part of input information and a critical reward signal, which has been largely overlooked in previous research despite its significance in ensuring manufacturability and performance. Previous methods often only consider the HPWL metric, suffering from optimizing different metrics effectively. By integrating regularity into the RL framework, our approach aligns more closely with advanced chip requirements.

The effectiveness of the proposed MaskRegulate is comprehensively evaluated on the ICCAD 2015 benchmark~\citep{iccad15}, which is is currently one of the largest open-source benchmarks that allows us to evaluate PPA metrics such as congestion and timing slack. We first compare the global HPWL and regularity of our approach against several competitive methods. Additionally, we use the commercial electronic design automation (EDA) tool \textit{Cadence Innovus} to evaluate the PPA performance, demonstrating that our proposed MaskRegulate can lead to significant PPA improvements, e.g., the placement layouts and two PPA metrics on superblue1, as shown in Figure~\ref{fig:congestion-fig1}. Specifically, compared to MaskPlace (an advanced RL placer~\cite{lai2022maskplace}; MaskRegulate shares a similar architecture to it), MaskRegulate improves 17.08\% on routing wirelength, 73.08\% and 38.81 \% on routed horizontal and vertical congestion overflow respectively, 18.35\% on worst negative slack, 37.89\% on total negative slack, and 46.17\% on the number of violation points.  

This work provides a more effective approach for macro placement of modern chips, opening new possibilities for the application of RL in chip design. The contributions of this work are highlighted in three key points:
\begin{itemize}
    \item \textbf{Novel problem formulation}: Innovatively applying RL in the refinement stage of macro placement, which allows for more effective learning from structured state and accurate reward information, significantly enhancing the learning efficiency and effectiveness.
    \item \textbf{Integration of regularity}: Introducing regularity, a critical yet previously overlooked metric in chip design, into the RL training framework, which not only aligns with industry practice but also enhances the chip PPA quality.
    \item \textbf{Impressive PPA improvement and comprehensive analysis}: On the popular ICCAD 2015 benchmark, our proposed MaskRegulate demonstrates significant improvements in PPA metrics, showing the practical applicability and effectiveness of the RL regulator.
\end{itemize}

\section{Background}

\subsection{Placement}\label{sec2.1-placement}
The circuit in the placement stage is considered as a graph where vertices model gates. The main input information is the netlist $\mathcal {N}=(V,E)$, where $V$ denotes the information (i.e., height and width) about all macros designated for placement on the chip, and $E$ is a hyper-graph comprised of nets $e_i\in E$, which encompasses multiple cells (including both macros and standard cells) and denotes their inter-connectivity in the routing stage. Given a netlist, a fixed canvas layout and a standard cell library, a placement method is expected to determine the appropriate physical locations of movable macros such that the total wirelength can be minimized. A macro placement solution $\bm{s}=\{(x_1,y_1), \dots, (x_k, y_k)\}$ consists of the positions of all the macros $\{v_i\}_{i=1}^k$, where $k$ denotes the total number of macros. One popular objective of macro placement is to minimize the total HPWL of all the nets while satisfying the cell density constraint, which is formulated as,
\begin{equation}\label{eq:obj}
\min_{\bm{s}} HPWL(\bm{s}) = \min_{\bm{s}} \sum_{e\in E}HPWL_e(\bm{s}), \ 
\text{s.t.} \ D(\bm{s}) \leq \epsilon,
\end{equation}
where $D$ denotes the density, $\epsilon$ is a threshold, and $HPWL_e$ is the HPWL of net $e$, which is defined as:
$ HPWL_e(\bm{s}) = (\max\nolimits_{v_i\in e} x_i - \min\nolimits_{v_i\in e} x_i) + (\max\nolimits_{v_i\in e} y_i - \min\nolimits_{v_i\in e} y_i)$. 

There are three mainstream placement methods, i.e., analytical methods, black-box optimization methods, and learning-based methods. Analytical methods~\cite{essential-issues-in-analytical} place macros and standard cells simultaneously, which can be roughly categorized into quadratic placement and nonlinear placement. Quadratic placement~\cite{ripple,polar} iterates between an unconstrained quadratic programming phase to minimize wirelength and a heuristic spreading phase to remove overlaps. Nonlinear placement~\cite{chen2008ntuplace3,lu2015eplace,cheng2018replace} formulates a nonlinear optimization problem and tries to directly solve it with gradient descent methods. Generally speaking, nonlinear placement can achieve better solution quality, while quadratic placement is more efficient. Recently, there has been extensive attention on GPU-accelerated non-linear placement methods. For example, DREAMPlace~\cite{lin2020dreamplace,dreamplace4} transforms the non-linear placement problem in Eq.~\eqref{eq:obj} into a neural network training problem, solves it by classical gradient descent and leverages GPU, enabling ultra-high parallelism and acceleration and producing state-of-the-art analytical placement quality. 

Black-box optimization methods for placement have a long history. Earlier methods such as SP~\cite{murata1996vlsi} and B$^*$-tree~\cite{chang2000b} have poor scalability due to the rectangular packing formulation. Recently, some black-box optimization methods have made significant progress by changing the search space. 
AutoDMP~\cite{agnesina2023autodmp} improves DREAMPlace by using Bayesian optimization to explore the configuration space and shows remarkable performance on multiple benchmarks. 
WireMask-BBO~\cite{wiremask-bbo} adopts a wire-mask-guided greedy genotype-phenotype mapping and can be equipped with any BBO algorithm, demonstrating the superior performance over other types of methods.

\subsection{RL for Macro Placement}\label{sec2.2}
Researchers recently leverage RL-based methods for better placement quality to meet the demands of modern chip design. GraphPlace~\cite{nature-graph} first models macro placement as a RL problem. It divides the chip canvas into discrete grids, with each macro assigned discrete coordinates of grids, wherein the agent decides the placement of the current macro at each step. However, no reward is given until all the macros are placed, making the reward sparse and hard to learn. DeepPR~\cite{deeppr} and PRNet~\cite{cheng2022the} incorporate macro placement, standard cells placement, and routing to achieve better performance than GraphPlace, but may violate the non-overlap constraint. To address this issue, MaskPlace~\cite{lai2022maskplace} introduces a dense reward and uses a pixel-level visual representation for circuit modules, which can comprehensively capture the configurations of thousands of pins, enabling fast placement in a full action space on a large canvas size. MaskPlace has many attractive benefits that previous methods do not have, e.g., 0\% overlap, dense reward, and high training efficiency. ChiPFormer~\cite{lai2023chipformer} incorporates an offline learning decision transformer and focuses on improving the generalizability of placer. 
EfficientPlace~\citep{fast-place} integrates a global tree search algorithm to guide the optimization process, achieving remarkable placement quality within a short time.

However, current RL methods exhibit several shortcomings: 1) Placing from scratch provides insufficient state information and inaccurate reward signals; 2) Most methods
focus on minimizing wirelength, which may bring macro blockages and thus harm the final PPA metrics. In this work, we propose a novel RL approach for macro placement: an RL policy acts as a macro regulator rather than a macro placer. Specifically, our learned RL policy is designed to adjust macros based on an existing placement result, rather than placing all macros from scratch. This approach aims to refine and optimize pre-existing layouts, addressing the limitations of traditional RL-based placement methods.

\section{Method}

We present our proposed MaskRegulate here. Section~\ref{sec:3.1} introduces our problem formulation and policy architecture, and Section~\ref{sec:3.2} describes how to integrate regularity into the method.

\subsection{MaskRegulate Framework}\label{sec:3.1}

\textbf{Problem formulation of RL regulator.} In the Markov Decision Process (MDP) formulation of traditional RL placer, a macro is placed at each step~\citep{nature-graph,deeppr,lai2022maskplace,lai2023chipformer}. 
The placement order of macros is determined based on some pre-defined rules, such as the number of nets, the size of macros, and the number of connected modules that have been placed.
An episode ends after all macros have been placed. Typically, the state representation includes information about the chip canvas, the macros that have already been placed, and the macro currently being placed. In GraphPlace~\citep{nature-graph}, the reward is determined only after all macros have been placed, resulting in a sparse reward signal that complicates the training process. Recent works have introduced various methods to densify the reward signal. For instance, WireMask~\citep{lai2022maskplace} provides a more continuous reward based on the macros already placed. In contrast to RL placers, our RL regulator focuses on refining an existing placement by adjusting the location of one macro at each step. Unlike the placer, which initiates the placement process from scratch, the regulator benefits from additional information when adjusting each macro. Specifically, the regulator considers not only the macros that have already been placed but also the positions of all other macros. Furthermore, it enhances accuracy by taking into account all macros, even while employing a reward function similar to WireMask.

Due to the advantages mentioned above in the MDP problem formulation, even without considering additional factors (e.g., regularity), RL regulator is able to achieve better results compared to RL placer, as shown in our experiments in Appendix~\ref{app:exp-ablation}. Furthermore, our main experimental results demonstrate superior performance not only in proxy metrics but also in PPA metrics measured by commercial tools, as shown in Section~\ref{sec:exp-rq1}. The regulator also exhibits better generalization abilities, as shown in Section~\ref{sec:exp-rq2}. Intuitively, adjusting an unseen chip is easier for the regulator compared to placing macros from scratch, as the incomplete state information of placer would be even worse in the case of unseen chips, resulting in poorer performance.

\begin{figure*}[t!]
\centering
\includegraphics[width=\textwidth]{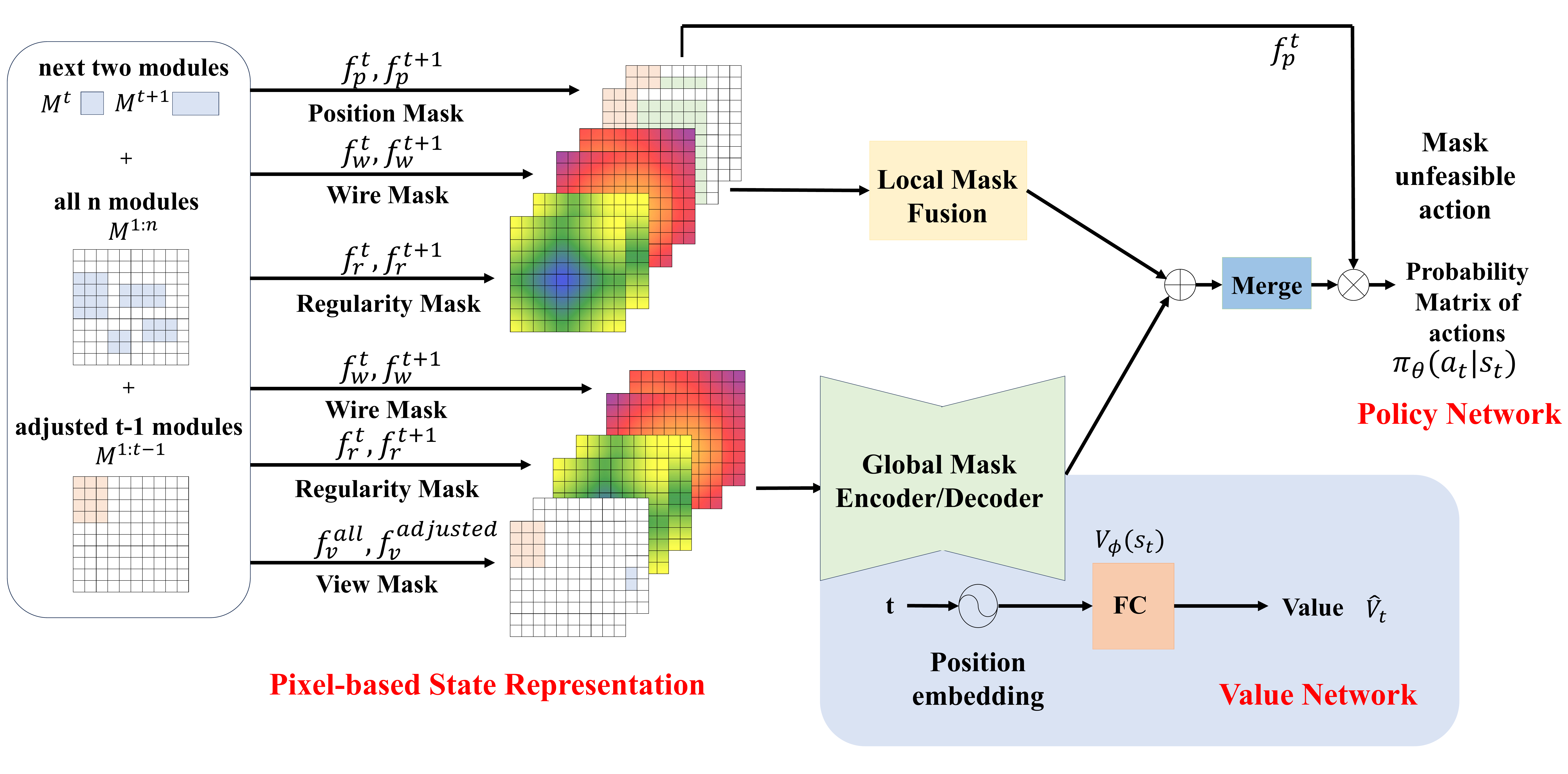}
\caption{Overview of MaskRegulate. MaskRegulate shares a similar architecture to MaskPlace~\citep{lai2022maskplace}, except for the MDP formulation and the integration of regularity in the state and reward.}
\label{fig:maskregulate-overview}
\end{figure*}

\begin{figure*}[t!]
\centering
\includegraphics[width=0.24\textwidth]{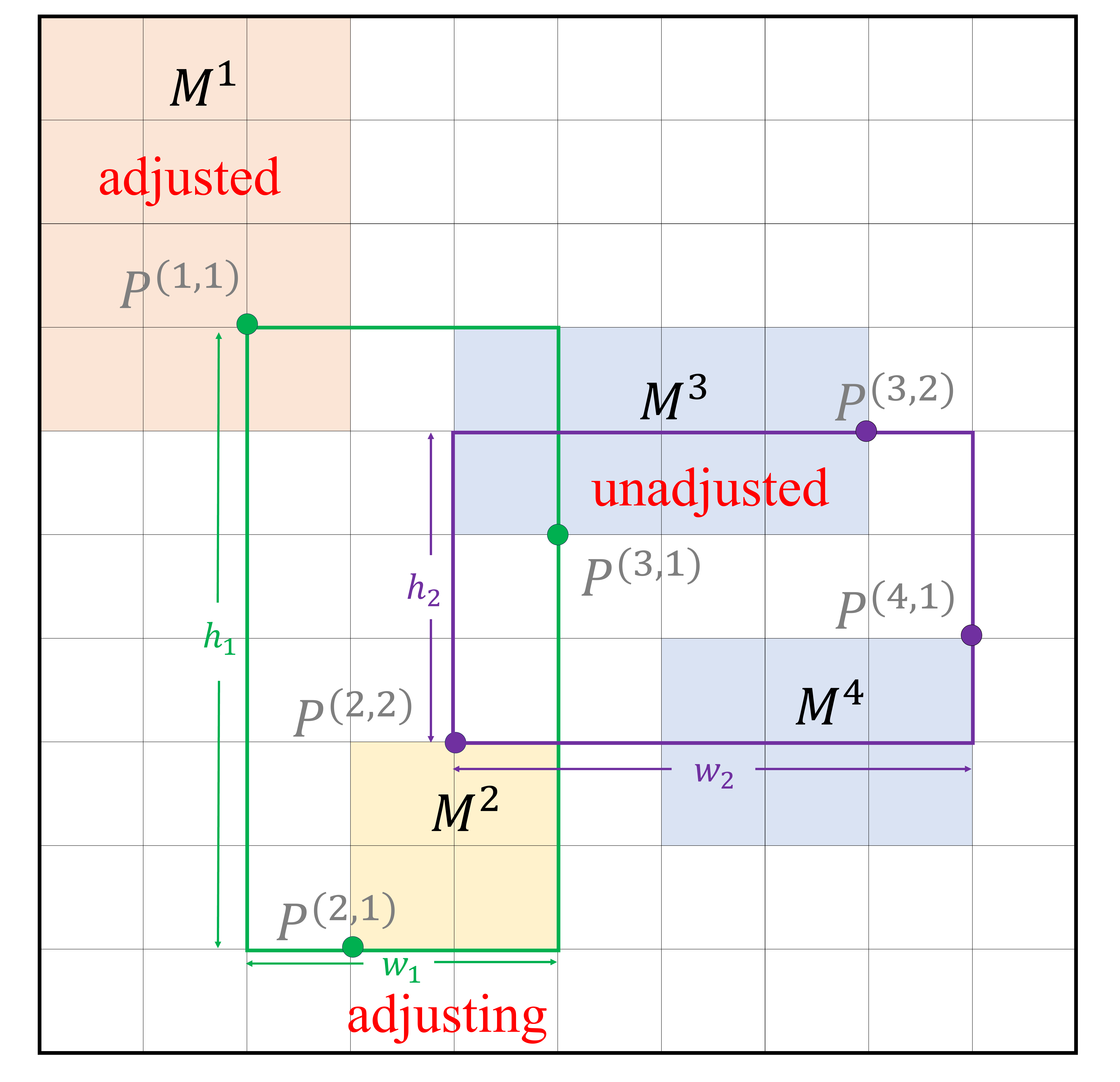} 
\includegraphics[width=0.24\textwidth]{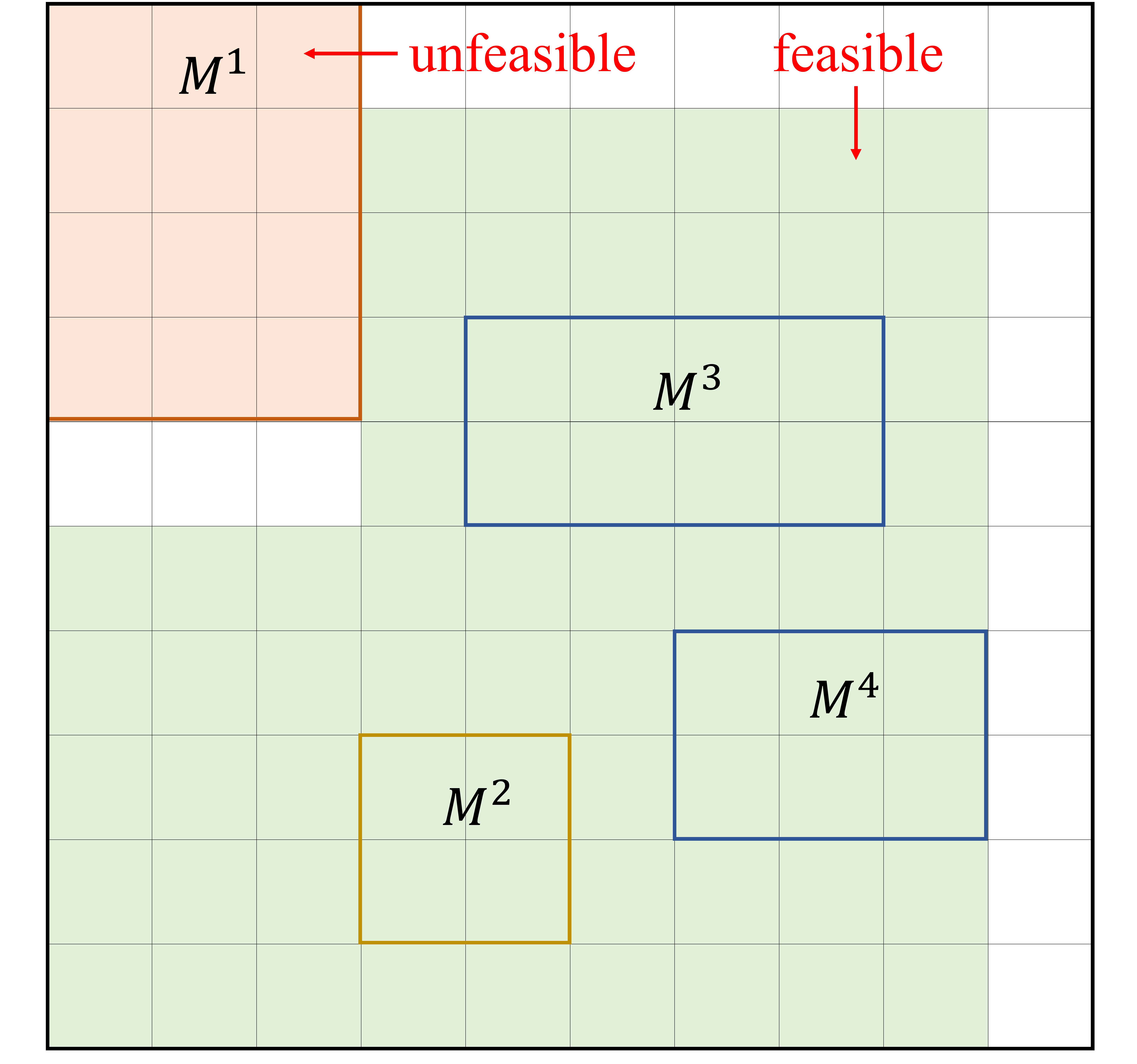} 
\includegraphics[width=0.24\textwidth]{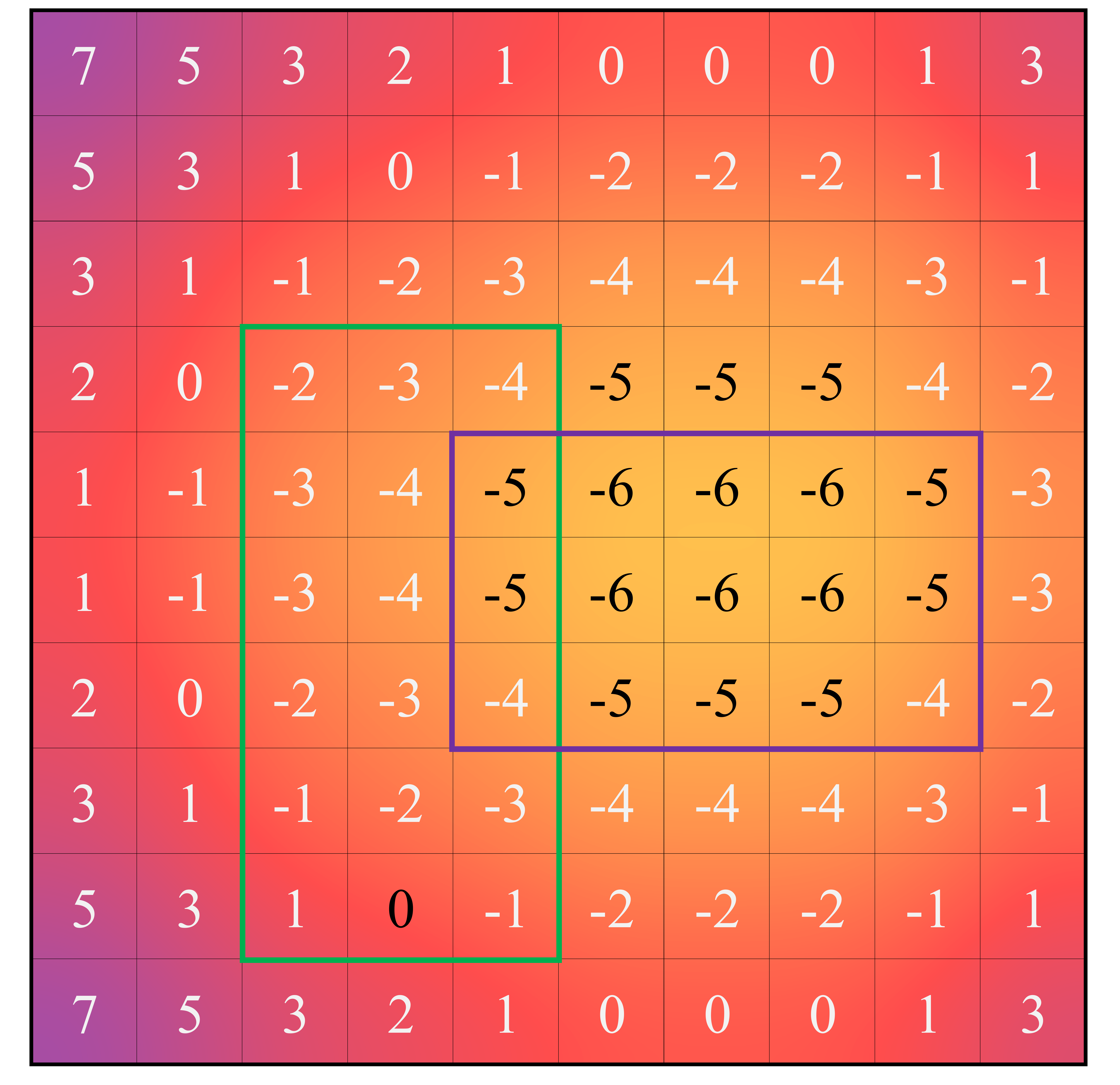} 
\includegraphics[width=0.24\textwidth]{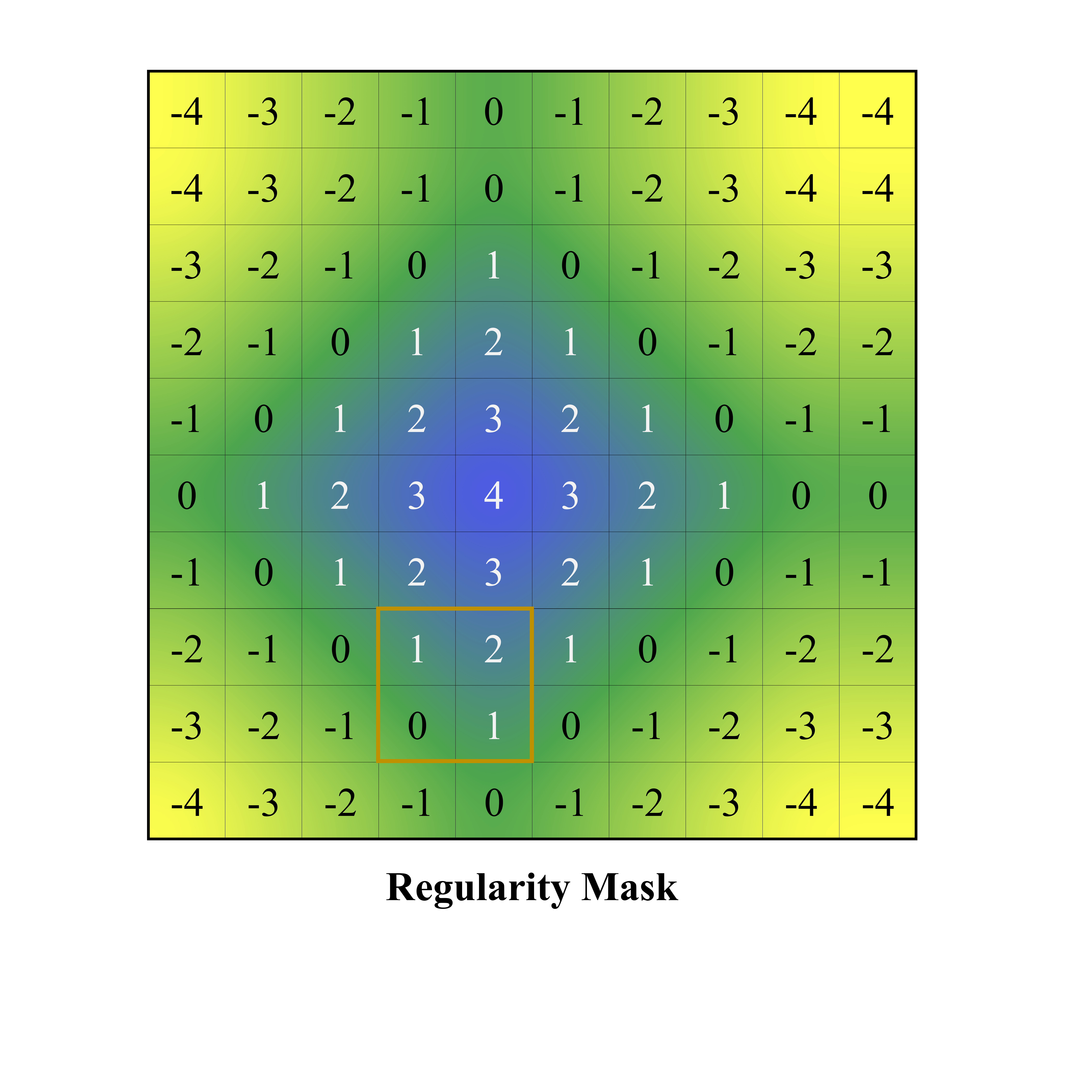} \\
\begin{minipage}{0.24\textwidth}
\centering
\textbf{(a) Chip canvas}
\end{minipage}
\begin{minipage}{0.24\textwidth}
\centering
\textbf{(b) PositionMask}
\end{minipage}
\begin{minipage}{0.24\textwidth}
\centering
\textbf{(c) WireMask}
\end{minipage}
\begin{minipage}{0.24\textwidth}
\centering
\textbf{(d) RegularMask}
\end{minipage}
\caption{Illustration of chip canvas, PositionMask, WireMask and RegularMask. We use the left-bottom corner of the module to denotes its location.}
\label{fig:illustration-mask}
\end{figure*}

\textbf{Policy architecture.} 
Our policy architecture is illustrated in Figure~\ref{fig:maskregulate-overview}. The policy divides the chip canvas into several grids and utilizes visual information as inputs, converting chip information into pixel-level image masks. This approach has demonstrated superior efficiency and performance in RL placer policy learning~\citep{lai2022maskplace,lai2023chipformer}. The inputs include an image of the current canvas, a PositionMask that identifies all valid positions for placing the current macro, a WireMask~\citep{lai2022maskplace} that indicates the approximate wirelength change for placing the current macro at each valid position, and a RegularMask that indicates the change in regularity for placing the current macro at each valid position (which will be detailed in Section~\ref{sec:3.2}). An illustration of the PositionMask and WireMask is provided in Figure~\ref{fig:illustration-mask}. To facilitate broader adjustments, the PositionMask has been modified to consider only macros that have already been adjusted; thus, grids occupied by unadjusted macros are available for placement. In our MaskRegulate, the calculation of the WireMask is based on all macros, allowing its value to either increase or decrease. These values are normalized to the range $[-1, 1]$, unlike the  $[0, 1]$ normalization used in \citep{lai2022maskplace}. 
Additionally, our framework introduces the RegularMask to quantify changes in regularity within the state and to encourage improvements in regularity through the reward function, as presented in Section~\ref{sec:3.2}.

\subsection{Integration of Regularity}\label{sec:3.2}

\textbf{Why does regularity matters?} 
Macro placement has significant impact on subsequent chip design processes, including standard cell placement and routing. If only focusing on minimizing wirelength (which is the case for most current RL placers), certain macros may end up positioned in the middle of the chip canvas, resulting in macro blockages~\citep{incre-macro}. This, in turn, leads to the division of available placement areas into separate and disconnected sub-regions. As a consequence, standard cells that are connected by the same net may be scattered across different placement sub-regions, resulting in increased overall wirelength and the potential routing challenges, which ultimately degrade the timing performance. Thus, a well-established practice among experienced engineers in macro placement is to place macros towards the peripheral regions of the chip to prevent macro blockage. In this work, we aim to integrate regularity in the learning-based placement approach to achieve placement preferences similar to those of experienced engineers.

\textbf{RegularMask.} Intuitively, macros closer to the edges tend to have lower regularity. Therefore, we propose a simple and effective way to measure regularity. On a canvas, the regularity of a grid located at $(x, y)$ is calculated as $\text{min} \{x, X_{\text{max}} - x \}$ + $\text{min} \{y, Y_{\text{max}}- y \}$, where $X_{\text{max}}$ and $Y_{\text{max}}$ represent the real length of the horizontal and vertical axes, respectively. Given a macro to be placed, the RegularMask measures the value change in regularity for each valid placement position, as illustrated in Figure~\ref{fig:illustration-mask}(d).

\textbf{Reward and policy learning.} 
The reward of MaskRegulate consists of two components: $r_{wire}$ and $r_{reg}$, which represent the reduction of HPWL and the improvement in regularity, respectively, after refining the current macro. To mitigate the influence of scale differences on training caused by wirelength and regularity, both $r_{wire}$ and $r_{reg}$ are normalized to $[0, 1]$. The final reward is $r = \alpha \cdot r_{wire} + (1-\alpha) \cdot r_{reg}$, where $\alpha$ is a trade-off coefficient. We will analyze the influence of $\alpha$ in Section~\ref{sec:exp-rq3}, showing that different $\alpha$ lead to different multi-objective preferences. The detailed information are presented in Appendix~\ref{app:methods}. MaskRegulate treats the chip canvas as a grid and divides it into $N \times N$ cells, resulting in $N^2$ possible discrete actions. We use the popular proximal policy optimization (PPO) algorithm~\citep{schulman2017proximal} to learn the regulator policy.

\section{Experiment}
In this section, we first introduce the basic experimental settings, including the tasks and evaluation metrics in Section~\ref{sec:exp-settings}. Then, we try to answer the following three research questions (RQs) in Sections~\ref{sec:exp-rq1} to~\ref{sec:exp-rq3}: 1) How does MaskRegulate perform compared to other methods? 2) How is the generalization ability of MaskRegulate? 3) How do the different parts of MaskRegulate affect the performance? Finally, we provide the visualization of placement results and congestion in Section~\ref{sec:vis}.

\subsection{Experimental Settings}\label{sec:exp-settings}

\textbf{Tasks.} We mainly use the ICCAD 2015 benchmark~\citep{iccad15} as our test-bed, which includes sufficient advanced chip information and is currently one of the largest open-source benchmarks that allows us to evaluate congestion, timing and other PPA metrics. The benchmark statistics are listed in Table~\ref{tab:iccad15} in Appendix~\ref{app:benchmarks}. Although ICCAD 2015 is the benchmark we have found that closely reflects the current practices in the EDA industry, it still has some shortcomings. For example, it allows for a large placement area, resulting in loose placement results that do not adhere to the design principles of advanced modern chips. Note that the ``A'' in PPA denotes ``Area'', which is a core metric of chip design and should be minimized~\cite{chan1991macro,tang2007memetic}. Therefore, we scale down the chip's placement area, presenting further challenges for the compared methods. Besides, we also conduct experiments on ISPD 2005 benchmark~\citep{nam2005ispd2005}, which is also a popular benchmark in AI for chip design but does not have sufficient information for PPA evaluation. Detailed results can be found in Appendix~\ref{app:additional-results}.

\textbf{Proxy evaluation metrics.}
We use the following two popular proxy metrics for a quick comparison of different algorithms: 1) Global HPWL. After determining the locations of all the macros, we use DREAMPlace~\citep{lin2020dreamplace} to place standard cells to obtain the global placement result, and then report the global HPWL (i.e., full HPWL involving both macros and standard cells). Compared to macro HPWL, global HPWL considers the total wirelength, typically on a scale that is two orders of magnitude larger, providing a better estimation of the final real performance of the chip. 2) Regularity: We compute the regularity values for all macros, which serve as a measurement of the overall regularity of the placement result. We run each algorithm for five times and report their mean and variance. We do not consider the rectangular uniform wire density (RUDY) metric~\citep{spindler2007fast} for congestion proxy, as this approximation is sometimes positively correlated with the HPWL metric and is not accurate~\citep{wiremask-bbo}. Instead, we will evaluate congestion within our PPA evaluation. 

\textbf{PPA evaluation metrics.}
The whole chip design process is lengthy and complex, and proxy metrics may not accurately capture the true performance of the chip. PPA metrics often require the use of commercial EDA tools to obtain precise results with expensive cost. In our experiments, we select the best placement result for PPA evaluation based on global HPWL from multiple runs. 
After obtaining the global placement results, we use commercial tool \textit{Cadence Innovus} to proceed the subsequent stages and evaluate their PPA metrics, including routed wirelength, routed vertical and horizontal congestion overflow, worst negative slack, total negative slack, and the number of violation points. These metrics are extremely important measures of chip design and are typically considered to evaluate the quality of a chip comprehensively.

\subsection{RQ1: How does MaskRegulate perform compared to other methods?}\label{sec:exp-rq1}

We consider the following methods to be compared: 
DREAMPlace~\citep{lin2020dreamplace}: A state-of-the-art analytical placer; 
AutoDMP~\citep{agnesina2023autodmp}: A method that improves DREAMPlace by exploring its configuration space iteratively;
WireMask-EA~\citep{wiremask-bbo}: A state-of-the-art black-box macro placement method with EA as the optimizer;
MaskPlace~\citep{lai2022maskplace}: A representative online RL methods, which shares similar policy architecture, state, HPWL reward with our MaskRegulate. 

For the same components, MaskPlace and MaskRegulate use the same settings, e.g., the number of grids, and the learning rate. Detailed information is provided in Appendix~\ref{app:methods}. Additionally, in order to demonstrate that the regulator has higher training efficiency than the placer, MaskRegulate and MaskPlace are trained for 1000 and 2000 episodes, respectively. For each chip, MaskRegulate uses DREAMPlace to obtain an initial macro placement result to be adjusted, which takes within few minutes and has relatively low quality. 

\begin{table*}[htbp]
\caption{Results of proxy metrics and PPA metrics on the ICCAD 2015 benchmarks.  
Global HPWL (1e8) and Regularity (1e6) are two proxy metrics.
PPA metrics are evaluated by \emph{Cadence Innovus}. The placement is performed by different methods, and the subsequent stages are performed by \emph{Cadence Innovus}. rWL (m) is the routed wirelength; rO-H (\%) and rO-V (\%) represent the routed horizontal and  vertical congestion overflow, respectively; WNS (ns) is the worst negative slack; TNS (1e5 $\mu$s) is the total negative slack; NVP (1e4) is the number of violation points. WNS and TNS are the larger the better, while the other metrics are the smaller the better. The best result of each metric on each chip is \textbf{bolded}.}
\resizebox{\textwidth}{!}{
\begin{tabular}{c|c|cc|cccccc}
\toprule
\multirow{2}{*}{Benchmark}                  & \multirow{2}{*}{Method}       & \multicolumn{2}{c}{Proxy metrics} & \multicolumn{6}{c}{PPA metrics}                                  \\
                              &              & Global HPWL  & Regularity & rWL & rO-H & rO-V  & WNS   & TNS & NVP \\ \midrule
\multirow{5}{*}{superblue1}   & DMP          & 8.96 $\pm$ 0.84    & 4.15 $\pm$ 0.04  & 154.23 & 17.15    & 4.48     & -119.616 & -2.91       & 3.35     \\
                              & AutoDMP      & 8.13 $\pm$ 0.17    & 4.99 $\pm$ 0.08  & 185.60 & 20.99    & 5.73     & -124.572 & -3.72       & 3.46     \\
                              & WireMask-EA & 8.07 $\pm$ 0.38    & 4.41 $\pm$ 0.15  & 149.49 & 7.62     & 0.38     & -67.616  & -3.57       & 2.94     \\
                              & MaskPlace    & 7.93 $\pm$ 0.06    & 4.40 $\pm$ 0.06  & 158.59 & 16.28    & 0.64     & -72.070  & -3.98       & 4.41     \\
                              & MaskRegulate & \textbf{5.77 $\pm$ 0.05}    & \textbf{3.31 $\pm$ 0.00}  & \textbf{116.11} & \textbf{1.26}     & \textbf{0.11}     & \textbf{-60.532}  & \textbf{-1.33}       & \textbf{1.06}     \\ \midrule
\multirow{5}{*}{superblue3}   & DMP          & 12.87 $\pm$ 1.73   & 4.43 $\pm$ 0.03  & 232.19 & 40.55    & 19.64    & -96.904  & -2.36       & 2.25     \\
                              & AutoDMP      & 8.13 $\pm$ 0.69    & 5.49 $\pm$ 0.17  & 166.15 & 14.71    & 3.39     & \textbf{-76.566}  & \textbf{-1.12}       & 1.44     \\
                              & WireMask-EA & 9.37 $\pm$ 0.81    & 4.77 $\pm$ 0.23  & 167.67 & 7.81     & 0.32     & -92.566  & -1.57       & 2        \\
                              & MaskPlace    & 8.90 $\pm$ 0.17    & 4.77 $\pm$ 0.06  & 177.25 & 9.16     & 0.64     & -111.041 & -1.77       & 2.02     \\
                              & MaskRegulate & \textbf{7.05 $\pm$ 0.03}    & \textbf{3.54 $\pm$ 0.00}  & \textbf{142.89} & \textbf{1.86}     & \textbf{0.18}     & -83.635  & -1.15       & \textbf{0.97}     \\ \midrule
\multirow{5}{*}{superblue4}   & DMP          & 6.81 $\pm$ 0.23    & 3.06 $\pm$ 0.01  & 132.16 & 20.62    & 4.87     & -73.192  & -1.63       & 2.42     \\
                              & AutoDMP      & 4.57 $\pm$ 0.78    & 3.41 $\pm$ 0.06  & 82.94  & 5.43     & 0.21     & \textbf{-48.137}  & \textbf{-0.64}       & 1.08     \\
                              & WireMask-EA & 5.51 $\pm$ 0.07    & 3.25 $\pm$ 0.10  & 110.20 & 8.29     & 0.61     & -83.233  & -1.85       & 1.98     \\
                              & MaskPlace    & 5.28 $\pm$ 0.03    & 3.22 $\pm$ 0.03  & 106.36 & 9.71     & 0.31     & -67.995  & -1.47       & 1.9      \\
                              & MaskRegulate & \textbf{4.15 $\pm$ 0.06}    & \textbf{2.18 $\pm$ 0.02}  & \textbf{81.78}  & \textbf{0.29}     & \textbf{0.11}     & -49.071  & -0.90       & \textbf{0.88}     \\ \midrule
\multirow{5}{*}{superblue5}   & DMP          & 8.78 $\pm$ 1.47    & 4.84 $\pm$ 0.06  & 144.64 & 3.75     & 0.46     & \textbf{-58.907}  & \textbf{-0.68}       & 1.64     \\
                              & AutoDMP      & 12.67 $\pm$ 4.09   & 5.79 $\pm$ 0.32  & 344.14 & 74.75    & 37.32    & -197.175 & -5.83       & 3.55     \\
                              & WireMask-EA & 10.23 $\pm$ 0.68   & 5.03 $\pm$ 0.15  & 189.84 & 4.06     & 0.41     & -75.115  & -1.83       & 2.18     \\
                              & MaskPlace    & 9.81 $\pm$ 0.03    & 4.86 $\pm$ 0.04  & 196.79 & 4.79     & 0.37     & -118.122 & -2.98       & 2.62     \\
                              & MaskRegulate & \textbf{6.94 $\pm$ 0.00}    & \textbf{4.23 $\pm$ 0.01}  & \textbf{137.79} & \textbf{0.02}     & \textbf{0.02}     & -74.83   & -0.73       & \textbf{1.32}     \\ \midrule
\multirow{5}{*}{superblue7}   & DMP          & 22.70 $\pm$ 0.91   & 4.24 $\pm$ 0.02  & 427.71 & 100.32   & 73.18    & -123.310 & -6.55       & 7.02     \\
                              & AutoDMP      & 10.04 $\pm$ 1.63   & 5.25 $\pm$ 0.09  & 221.69 & 6.32     & 0.64     & -52.556  & -1.82       & 4.31     \\
                              & WireMask-EA & 10.05 $\pm$ 0.34   & 4.31 $\pm$ 0.13  & 195.36 & 0.82     & 0.35     & -73.070  & -1.78       & 3.45     \\
                              & MaskPlace    & 9.99 $\pm$ 0.05    & 4.36 $\pm$ 0.03  & 204.32 & 2.77     & \textbf{0.33}     & -69.441  & -2.18       & 5.78     \\
                              & MaskRegulate & \textbf{7.90 $\pm$ 0.03}    & \textbf{3.03 $\pm$ 0.00}  & \textbf{162.72} & \textbf{0.59}     & 0.64     & \textbf{-50.494}  & \textbf{-1.48}       & \textbf{2.27}     \\ \midrule
\multirow{5}{*}{superblue10}  & DMP          & 14.81 $\pm$ 1.33   & 4.33 $\pm$ 0.02  & 261.35 & 7.00     & 5.16     & -83.509  & -3.09       & 2.51     \\
                              & AutoDMP      & 11.48 $\pm$ 1.78   & 6.55 $\pm$ 0.15  & 234.24 & 2.83     & 1.37     & -169.540 & -2.84       & 1.71     \\
                              & WireMask-EA & 13.52 $\pm$ 2.25   & 4.82 $\pm$ 0.12  & 223.08 & 0.76     & 0.55     & -121.785 & -3.11       & \textbf{1.65}     \\
                              & MaskPlace    & \textbf{10.94 $\pm$ 0.21}   & 4.75 $\pm$ 0.11  & 212.85 & \textbf{0.51}    & 0.19     & -81.916  & -4.37       & 2.11     \\
                              & MaskRegulate & 11.23 $\pm$ 0.38   & \textbf{3.90 $\pm$ 0.44}  & \textbf{212.84} & 0.52     & \textbf{0.09}     & \textbf{-77.980}  & \textbf{-2.74}       & 1.75     \\ \midrule
\multirow{5}{*}{superblue16}  & DMP          & 10.22 $\pm$ 0.58   & 2.78 $\pm$ 0.02  & 187.03 & 74.53    & 30.31    & -138.370 & -4.14       & 4.95     \\
                              & AutoDMP      & 6.12 $\pm$ 2.49    & 3.68 $\pm$ 0.09  & 119.62 & 2.46     & 0.22     & -41.292  & \textbf{-1.17}       & 2.26     \\
                              & WireMask-EA & 6.13 $\pm$ 0.14    & 3.17 $\pm$ 0.17  & 120.45 & 8.20     & 0.36     & -89.395  & -2.19       & 2.66     \\
                              & MaskPlace    & \textbf{5.53 $\pm$ 0.04}    & 3.08 $\pm$ 0.06  & 110.52 & 2.46     & \textbf{0.11}     & \textbf{-36.488}  & -1.36       & 2.51     \\
                              & MaskRegulate & 5.53 $\pm$ 0.04    & \textbf{2.55 $\pm$ 0.12}  & \textbf{106.82} & \textbf{1.40}     & 0.17     & -45.962  & -1.77       & \textbf{1.88}     \\ \midrule
\multirow{5}{*}{superblue18}  & DMP          & 4.97 $\pm$ 1.38    & 2.44 $\pm$ 0.00  & 85.93  & 11.76    & 8.93     & -73.429  & -0.46       & 0.95     \\
                              & AutoDMP      & 3.02 $\pm$ 0.01    & 3.01 $\pm$ 0.10  & 61.57  & 1.02     & \textbf{0.02}     & \textbf{-17.545}  & \textbf{-0.32}       & \textbf{0.69}     \\
                              & WireMask-EA & 3.46 $\pm$ 0.07    & 2.62 $\pm$ 0.09  & 69.25  & 1.04     & 0.27     & -34.143  & -0.43       & 1.43     \\
                              & MaskPlace    & 3.49 $\pm$ 0.06    & 2.59 $\pm$ 0.08  & 70.24  & 1.08     & 0.40     & -43.869  & -0.76       & 1.31     \\
                              & MaskRegulate & \textbf{2.96 $\pm$ 0.01}    & \textbf{1.55 $\pm$ 0.00}  & \textbf{60.64}  & \textbf{0.04}     & 0.03     & -28.285  & -0.40       & 0.85     \\ \midrule
\multirow{5}{*}{Average Rank} & DMP          & 4.625              & 2                & 4.375  & 4.5      & 4.75     & 3.875    & 3.625       & 4.125    \\
                              & AutoDMP      & 3                  & 5                & 3.375  & 3.5      & 3.375    & 2.75     & 2.25        & 2.5      \\
                              & WireMask-EA & 3.75               & 3.75             & 3      & 2.75     & 2.875    & 3.375    & 3.25        & 3        \\
                              & MaskPlace    & 2.375              & 3.25             & 3.125  & 3.125    & 2.375    & 3.125    & 4           & 3.875    \\
                              & MaskRegulate & \textbf{1.25}               & \textbf{1}                & \textbf{1.125}  & \textbf{1.125}    & \textbf{1.625}    & \textbf{1.875}    & \textbf{1.875}       & \textbf{1.5}     \\\bottomrule
\end{tabular}
}
\label{tab:ppa}
\end{table*}

The overall evaluation results are shown in Table~\ref{tab:ppa}.
MaskRegulate achieves the best average rank on both proxy and PPA metrics. 
DREAMPlace has the worst average ranking on wirelength, congestion, and timing. However, after adjustment by MaskRegulate, the obtained placements achieve the best average rank. Compared to MaskPlace, MaskRegulate leads to significant improvements in multiple PPA indicators:  improves 17.08\% on routing wirelength, 73.08\% and 38.81 \% on routed horizontal and vertical congestion overflow respectively, 18.35\% on worst negative slack, 37.89\% on total negative slack, and 46.17\% on the number of violation points. By incorporating regularity, MaskRegulate achieves the highest regularity on all the eight chips.
We can observe a certain correlation between the proxy metric (global HPWL) and the real metric (rWL), but there still exists a gap, indicating the challenges involved in the placement task.
Furthermore, we provide detailed visualizations of placement results in Figure~\ref{fig:congestions-all}, where MaskRegulate shows significant improvements on congestion metrics. Besides, the final placement layouts of MaskRegulate are much regular than all the other methods. 

\subsection{RQ2: How is the generalization ability of MaskRegulate?}\label{sec:exp-rq2}

The generalization ability of RL policies is an important question to be investigated. 
In this section, we pre-train MaskRegulate and MaskPlace on the first four chips (i.e., superblue1, superblue3, superblue4, and superblue5) and test on the remaining four chips. To further validate the ability of MaskRegulate to adjust different initial placement results, we use it to adjust the results obtained by different initial placements on the test chips. 

\begin{table}[htbp]
\caption{Generalization results of proxy metrics on the four chips of ICCAD 2015 benchmarks. The best result of each metric on each chip is \textbf{bolded}.}
\begin{tabular}{c|cc|cc}
\toprule
\multirow{2}{*}{Benchmark} & \multicolumn{2}{c}{MaskRegulate}            & \multicolumn{2}{|c}{MaskPlace} \\
                           & Global HPWL (1e8)                & Regularity   (1e6)       & Global HPWL (1e8)        & Regularity   (1e6) \\\midrule
superblue7                 & \textbf{7.99 $\pm$ 0.06}  & \textbf{3.04 $\pm$ 0.00} & 10.33 $\pm$ 0.17  & 4.24 $\pm$ 0.06    \\
superblue10                & \textbf{11.55 $\pm$ 0.27} & \textbf{3.25 $\pm$ 0.00} & 11.88 $\pm$ 0.72  & 4.73 $\pm$ 0.05    \\
superblue16                & \textbf{5.16 $\pm$ 0.05}  & \textbf{1.87 $\pm$ 0.00} & 5.97 $\pm$ 0.20   & 3.11 $\pm$ 0.03    \\
superblue18                & \textbf{3.04 $\pm$ 0.02}  & \textbf{1.54 $\pm$ 0.00} & 3.69 $\pm$ 0.09   & 2.52 $\pm$ 0.03   \\\bottomrule
\end{tabular}~\label{tab:generalization}
\end{table}
The results are shown in Table~\ref{tab:generalization}. On both the global HPWL and regularity metrics, MaskRegulate consistently outperforms MaskPlace, showcasing its stronger generalization capability. An interesting finding is that MaskRegulate performs better on unseen chips than on the chips it was trained on, specifically in terms of global HPWL, such as with superblue16. This may suggest that MaskRegulate has learned some general knowledge during the pre-training process, enabling it to overcome local optima that may arise from direct learning on the target chip.

\subsection{RQ3: How do the different parts of MaskRegulate affect the performance?}\label{sec:exp-rq3}
We investigate the influence of different parts and provide additional analysis in this section.

\begin{figure*}[t!]
\centering
\includegraphics[width=0.24\textwidth]{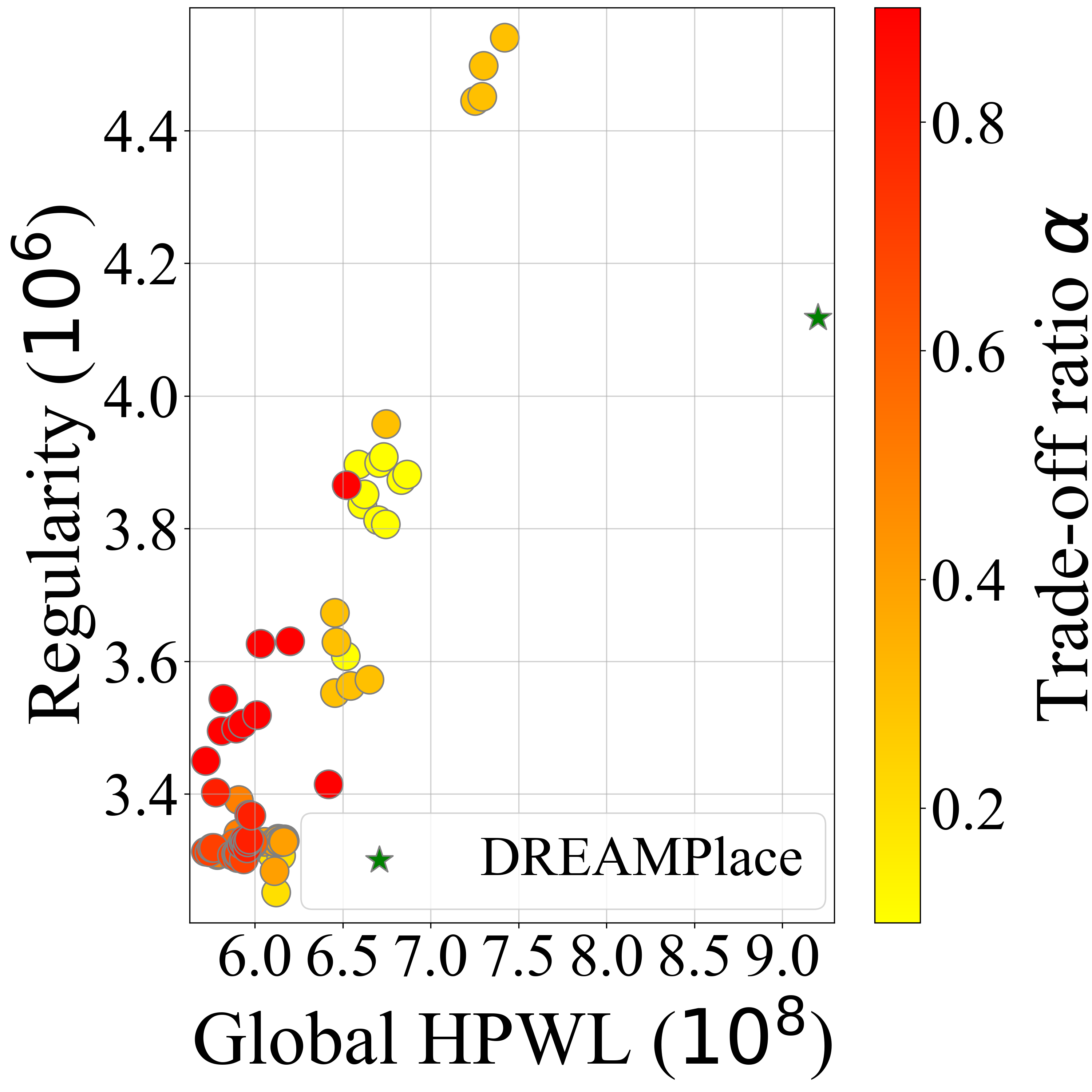} 
\includegraphics[width=0.24\textwidth]{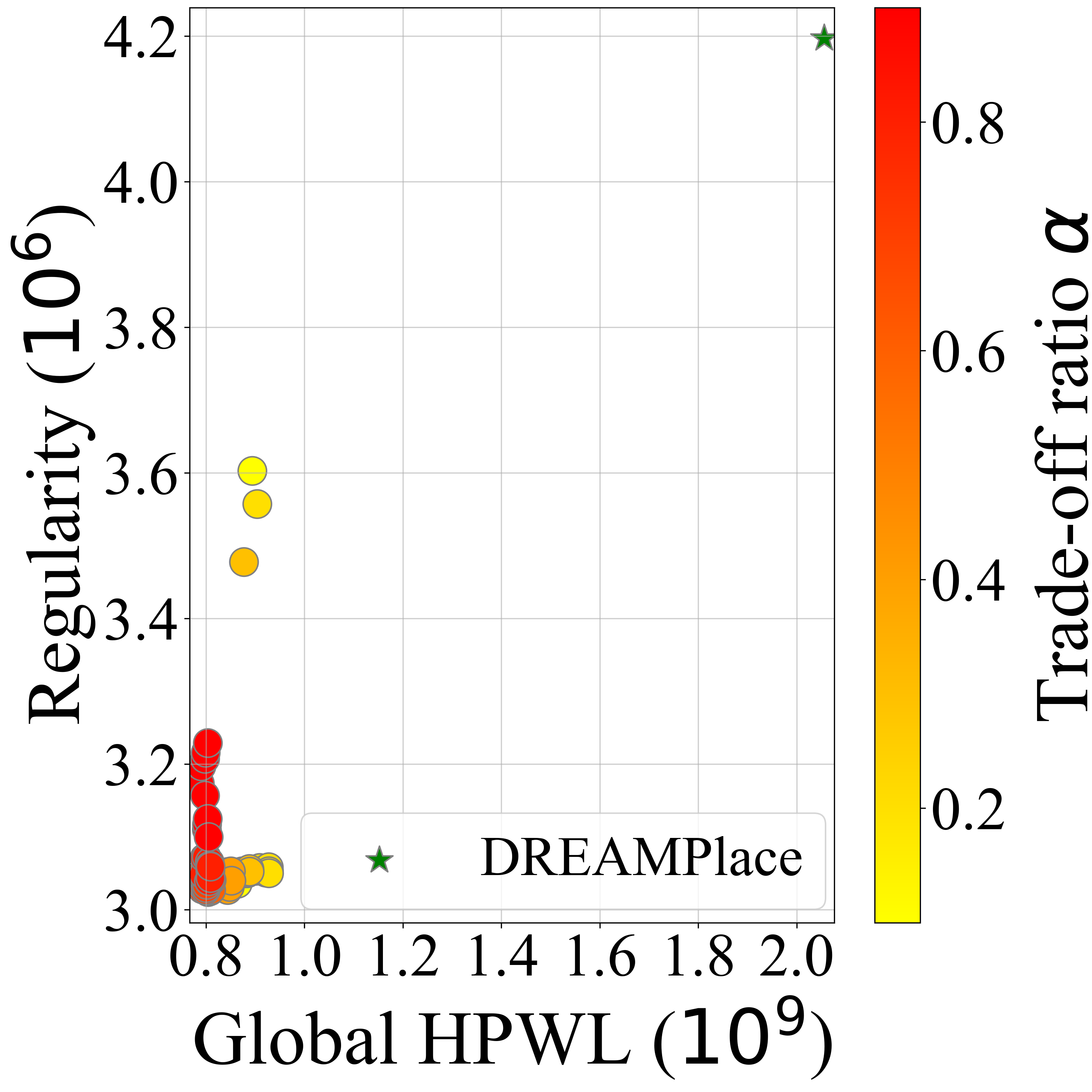} 
\includegraphics[width=0.24\textwidth]{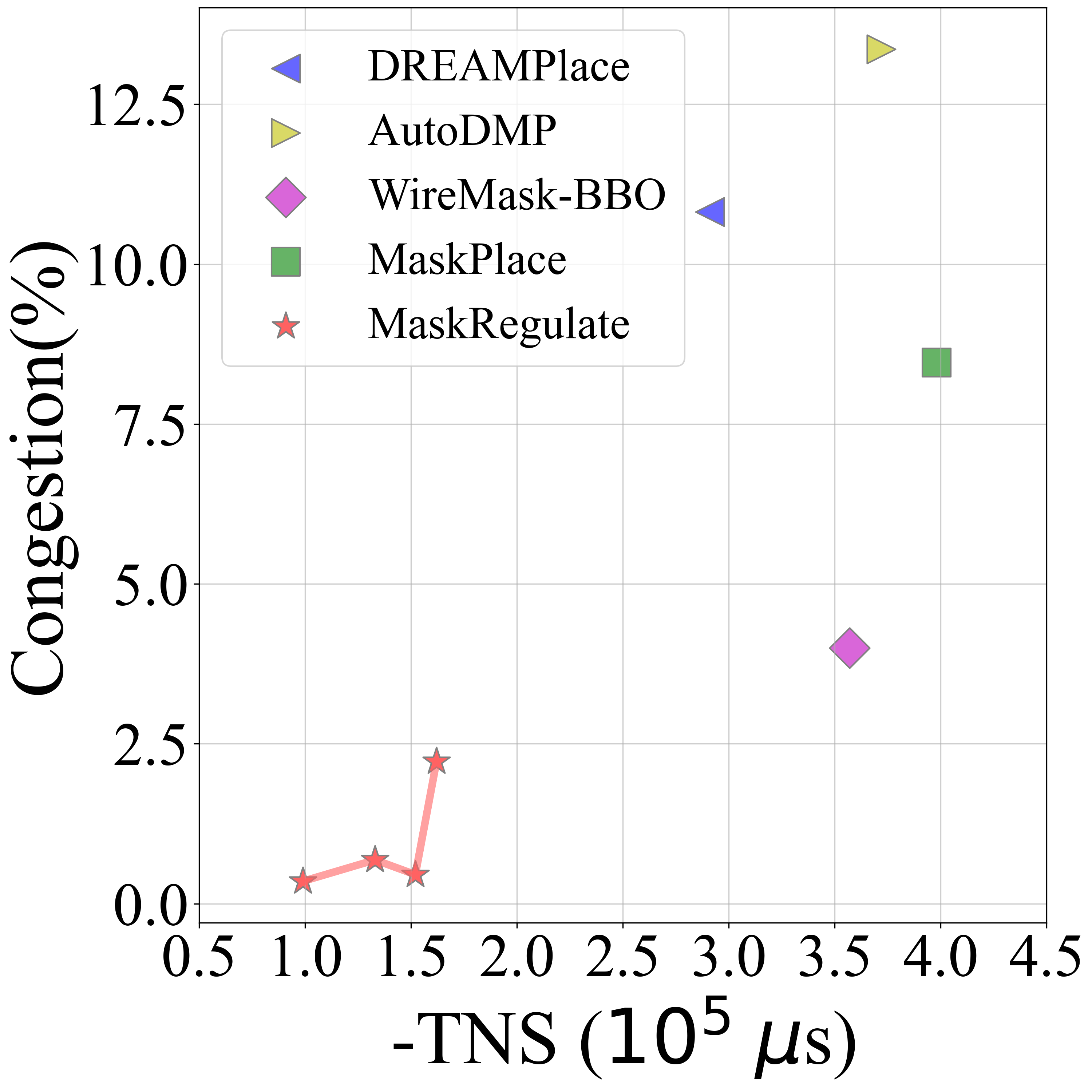} 
\includegraphics[width=0.24\textwidth]{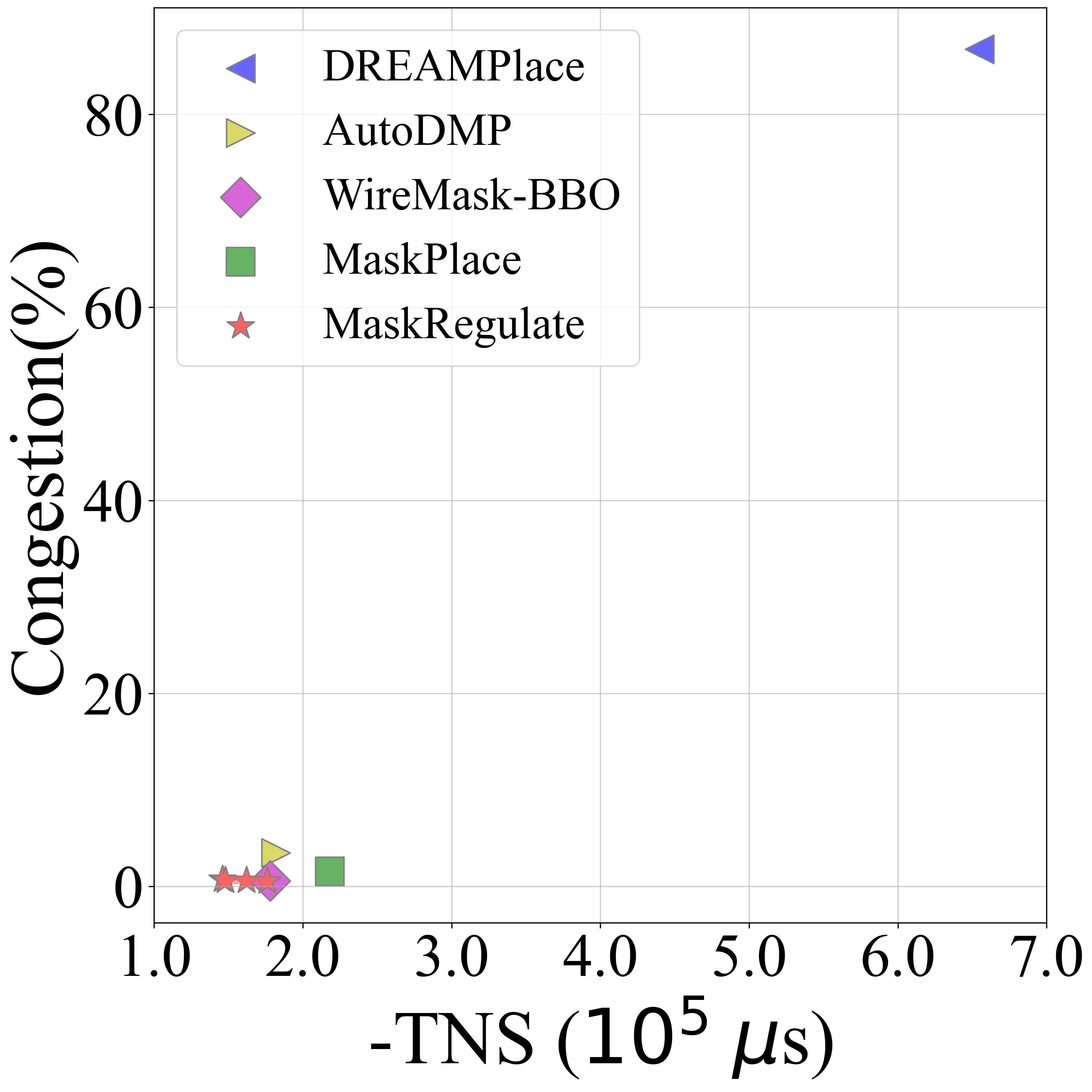} \\
\begin{minipage}{0.24\textwidth}
\centering
\textbf{\small (a) Proxy metrics on superblue1}
\end{minipage}
\begin{minipage}{0.24\textwidth}
\centering
\textbf{\small (b) Proxy metrics on superblue7}
\end{minipage}
\begin{minipage}{0.24\textwidth}
\centering
\textbf{\small (c) PPA metrics on superblue1}
\end{minipage}
\begin{minipage}{0.24\textwidth}
\centering
\textbf{\small (d) PPA metrics on superblue7}
\end{minipage}
\caption{Illustration of MaskRegulate regulators with varying $\alpha$ values (ranging from 0.1 to 0.9).}
\label{fig:mo-analysis}
\end{figure*}

\textbf{Hyperparameters sensitivity analysis: different trade-off coefficient $\alpha$ leads to different multi-objective preferences.}
One hyperparameter of RegularMask is the coefficient $\alpha$ between HPWL reward $r_{wire}$ and regularity reward $r_{reg}$, where a higher $\alpha$ indicates a preference for optimizing HPWL, and vice versa. In this section, we investigate the influence of the trade-off coefficient $\alpha$. We train different MaskRegulate regulators with varying $\alpha$ values (ranging from 0.1 to 0.9) and report the proxy and PPA results in Figure~\ref{fig:mo-analysis}. Due to the expensive computational cost of PPA, we select four different trade-offs of MaskRegulate for evaluation. As expected, different $\alpha$ values lead to different multi-objective preferences. In our experiments, we use $\alpha=0.7$ for all the chips as it achieves a relative balance between different objectives.

\textbf{Ablation studies.} 
We consider the following ablations of MaskRegulate. 
1) Only changing the problem formulation and purely comparing placer and regulator. We implement Vanilla-MaskRegulate, where the only difference to MaskPlace is the problem formulation, and all the other components (e.g., state and reward) are the same. The results show that Vanilla-MaskRegulate consistently outperforms MaskPlace in terms of Global HPWL. 
2) MaskRegulate with or without normalization. Since global HPWL has large scale than regularity, MaskRegulate w/o normalization does not prefer to consider regularity, which is not what we expect. 
3) Training regularity-aware RL placer from scratch. We implement MaskPlace + RegularMask and compare it with MaskPlace and MaskRegulate. The results show the advantages of the integration of regularity (between MaskPlace and MaskPlace + RegularMask) and our RL regular formulation (between MaskPlace + RegularMask and MaskRegulate). 
The above ablation results demonstrate the effectiveness of each component of MaskRegulate. Detailed results and discussions are provided in Appendix~\ref{app:exp-ablation} due to space limitation.

\subsection{Visualizations of placement results and congestion.}~\label{sec:vis}
\begin{figure*}[t!]
\includegraphics[width=0.18\textwidth, keepaspectratio]{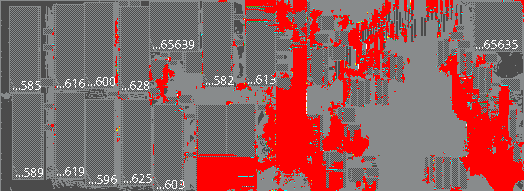}
\includegraphics[width=0.18\textwidth, keepaspectratio]{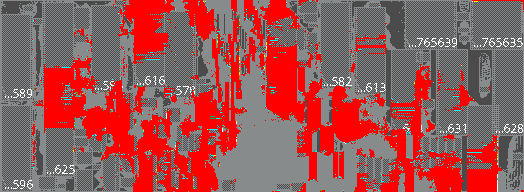}
\includegraphics[width=0.18\textwidth, keepaspectratio]{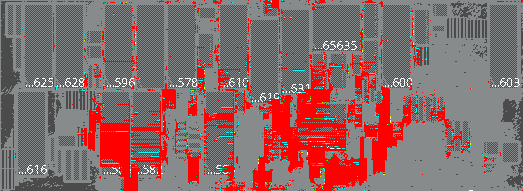}
\includegraphics[width=0.18\textwidth, keepaspectratio]{PPA/superblue1_MaskPlace.png}
\includegraphics[width=0.18\textwidth, keepaspectratio]{PPA/superblue1_MaskRegulate.png} \\
\includegraphics[width=0.18\textwidth, keepaspectratio]{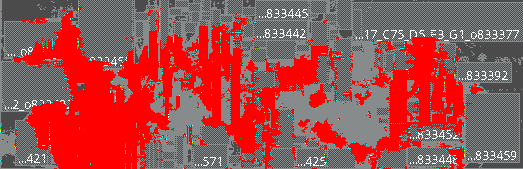}
\includegraphics[width=0.18\textwidth, keepaspectratio]{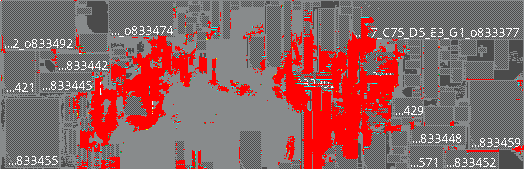}
\includegraphics[width=0.18\textwidth, keepaspectratio]{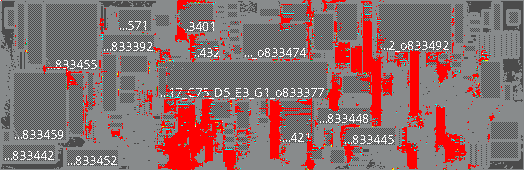}
\includegraphics[width=0.18\textwidth, keepaspectratio]{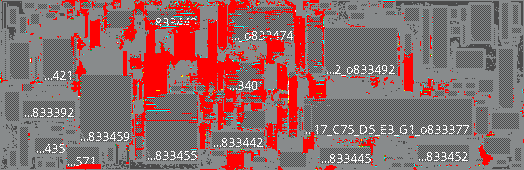}
\includegraphics[width=0.18\textwidth, keepaspectratio]{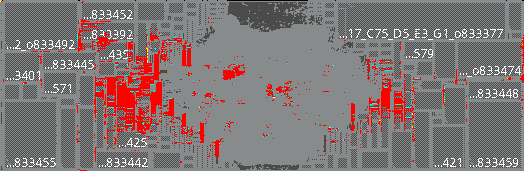} \\
\includegraphics[width=0.18\textwidth, keepaspectratio]{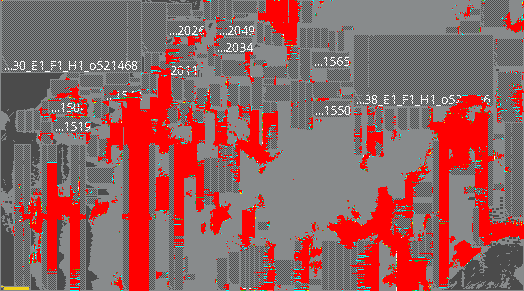}
\includegraphics[width=0.18\textwidth, keepaspectratio]{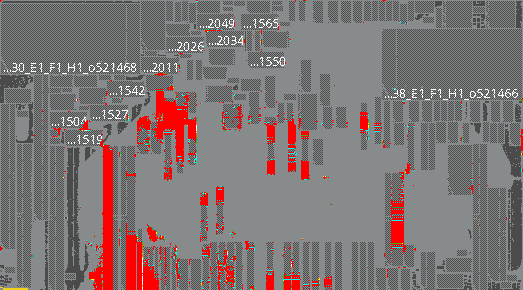}
\includegraphics[width=0.18\textwidth, keepaspectratio]{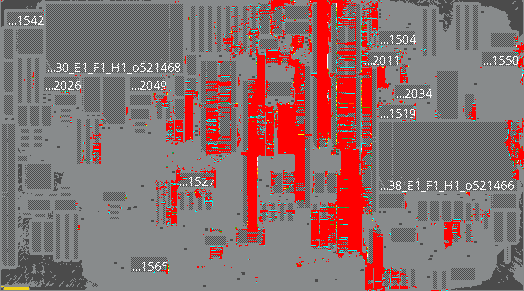}
\includegraphics[width=0.18\textwidth, keepaspectratio]{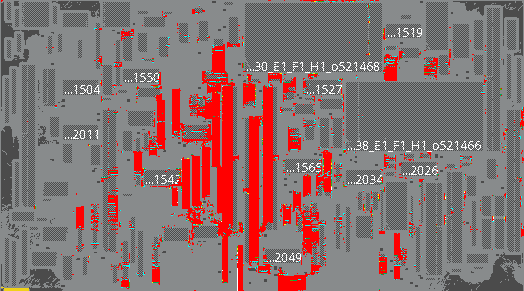}
\includegraphics[width=0.18\textwidth, keepaspectratio]{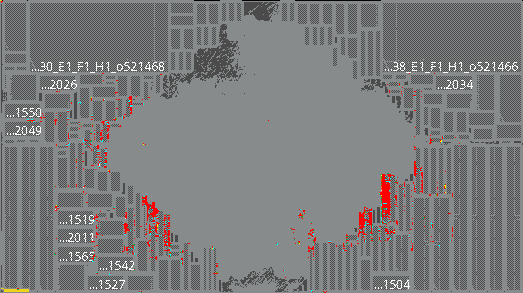} \\
\includegraphics[width=0.18\textwidth, keepaspectratio]{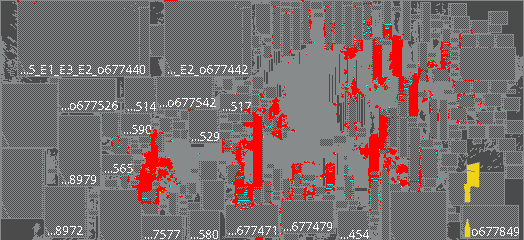}
\includegraphics[width=0.18\textwidth, keepaspectratio]{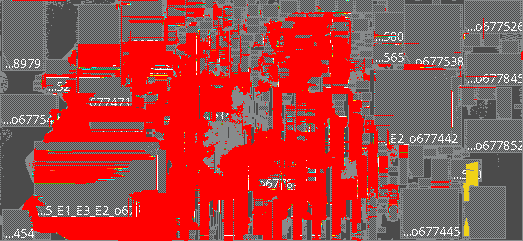}
\includegraphics[width=0.18\textwidth, keepaspectratio]{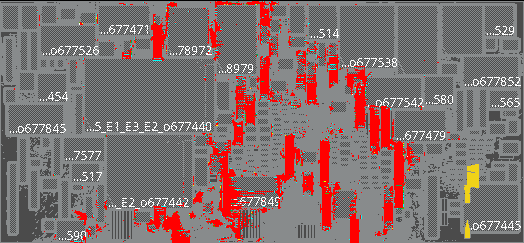}
\includegraphics[width=0.18\textwidth, keepaspectratio]{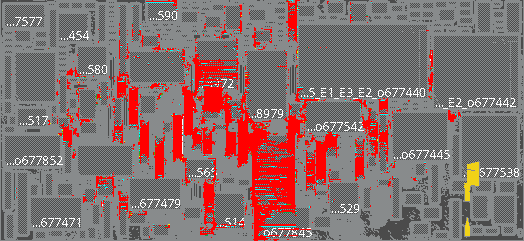}
\includegraphics[width=0.18\textwidth, keepaspectratio]{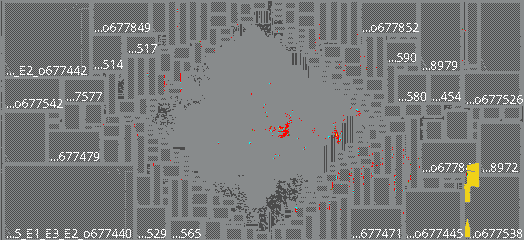} \\
\includegraphics[width=0.18\textwidth, keepaspectratio]{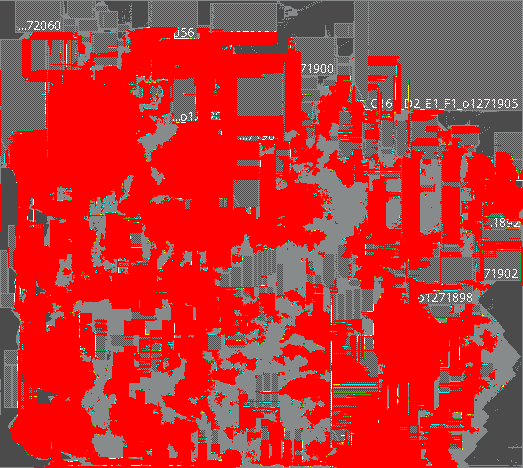}
\includegraphics[width=0.18\textwidth, keepaspectratio]{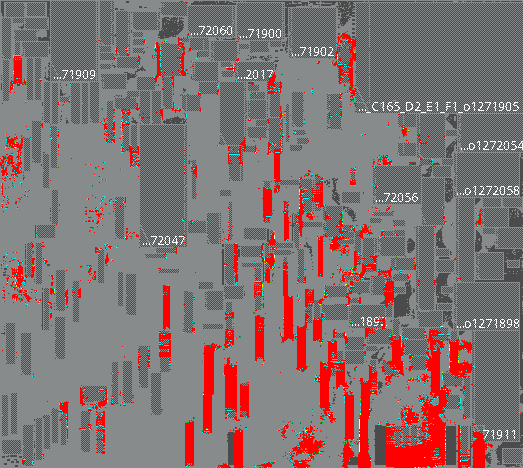}
\includegraphics[width=0.18\textwidth, keepaspectratio]{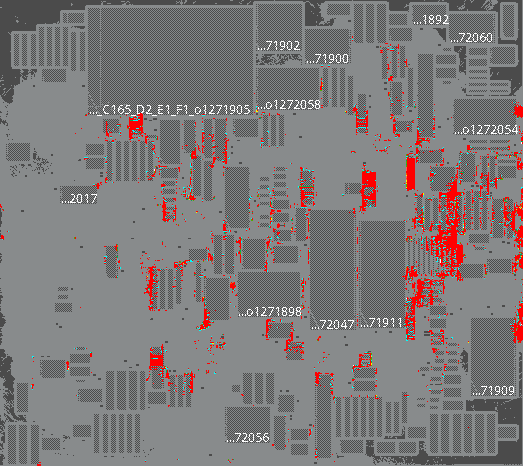}
\includegraphics[width=0.18\textwidth, keepaspectratio]{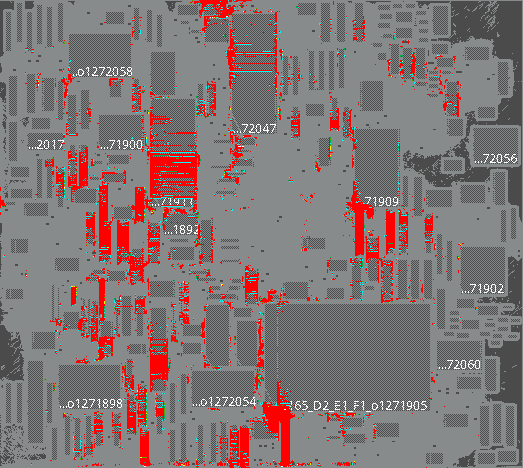}
\includegraphics[width=0.18\textwidth, keepaspectratio]{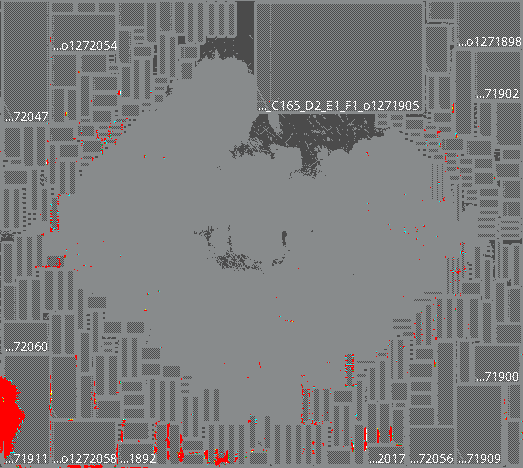} \\
\includegraphics[width=0.18\textwidth, keepaspectratio]{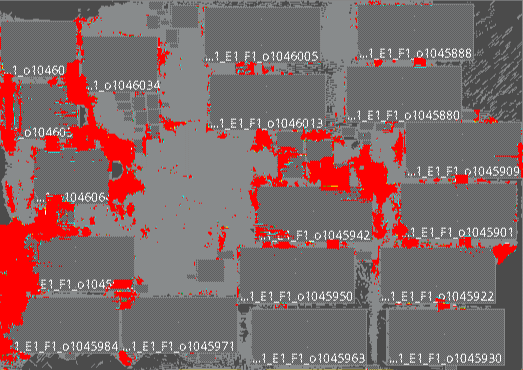}
\includegraphics[width=0.18\textwidth, keepaspectratio]{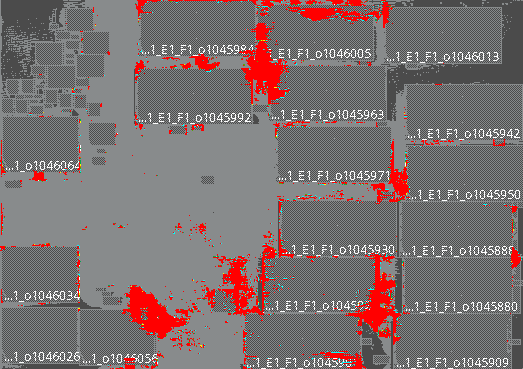}
\includegraphics[width=0.18\textwidth, keepaspectratio]{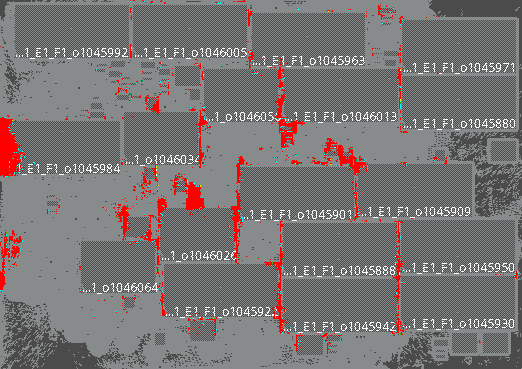}
\includegraphics[width=0.18\textwidth, keepaspectratio]{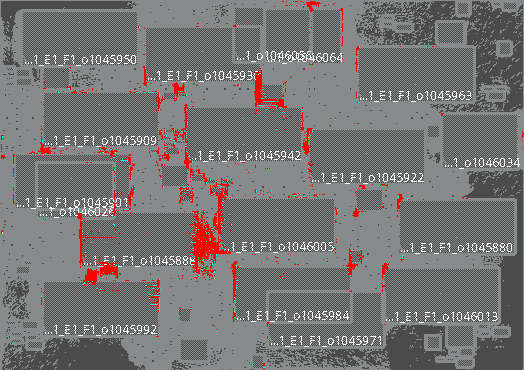}
\includegraphics[width=0.18\textwidth, keepaspectratio]{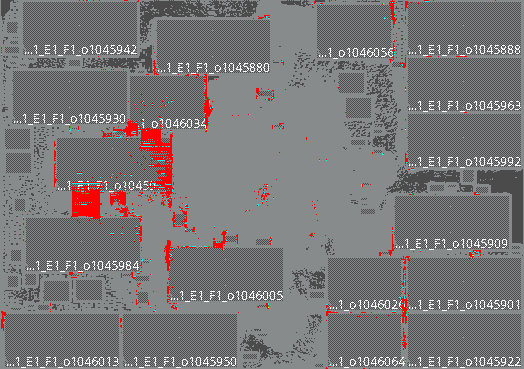} \\
\includegraphics[width=0.18\textwidth, keepaspectratio]{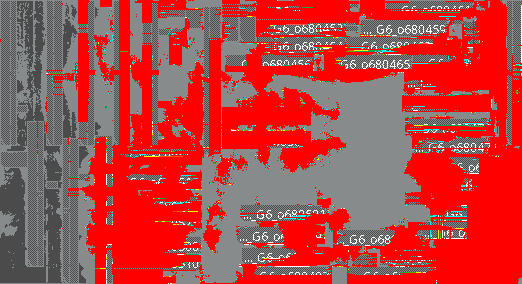}
\includegraphics[width=0.18\textwidth, keepaspectratio]{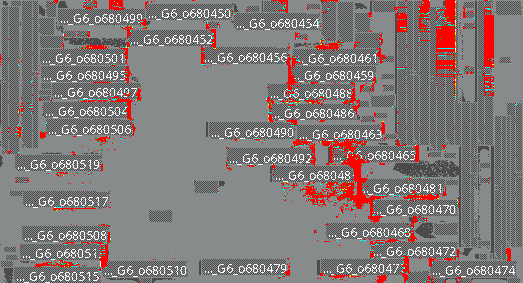}
\includegraphics[width=0.18\textwidth, keepaspectratio]{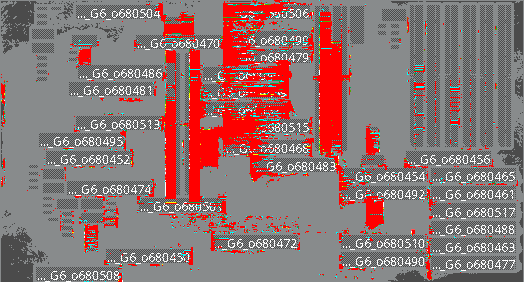}
\includegraphics[width=0.18\textwidth, keepaspectratio]{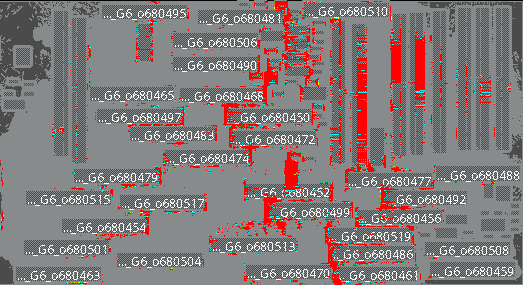}
\includegraphics[width=0.18\textwidth, keepaspectratio]{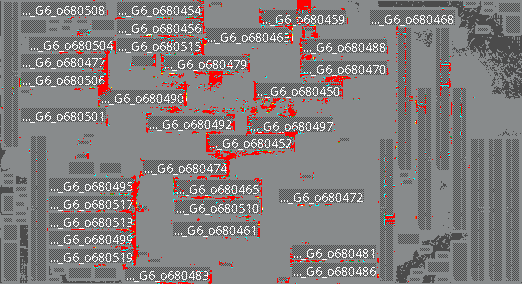} \\
\includegraphics[width=0.18\textwidth, keepaspectratio]{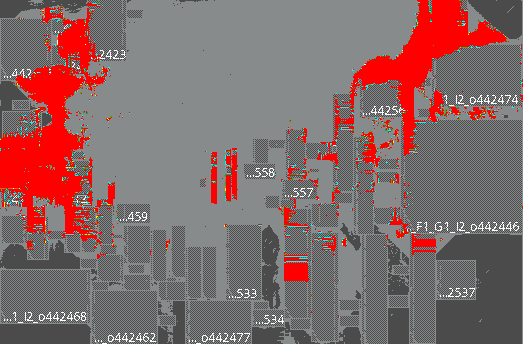}
\includegraphics[width=0.18\textwidth, keepaspectratio]{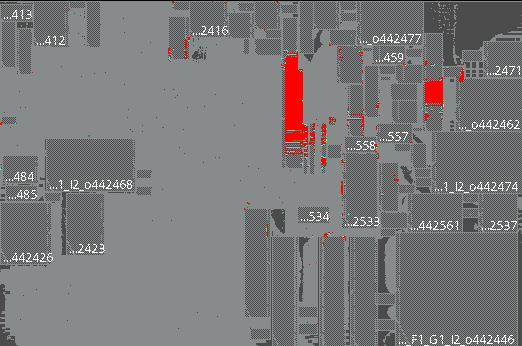}
\includegraphics[width=0.18\textwidth, keepaspectratio]{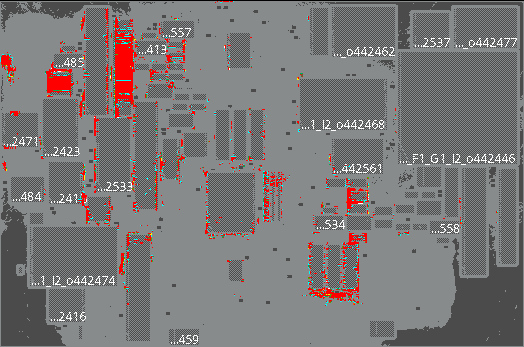}
\includegraphics[width=0.18\textwidth, keepaspectratio]{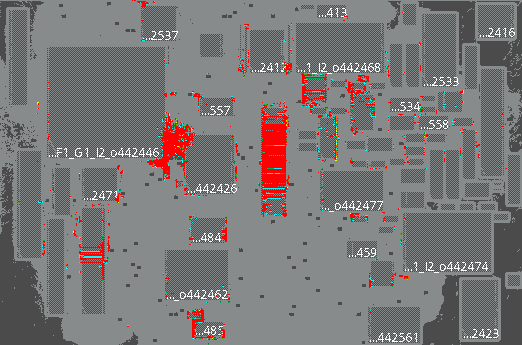}
\includegraphics[width=0.18\textwidth, keepaspectratio]{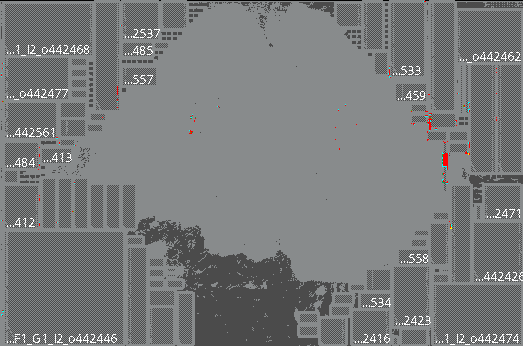} \\
\begin{minipage}{0.18\textwidth}
\centering
\textbf{ \small (a) DREAMPlace}
\end{minipage}
\begin{minipage}{0.18\textwidth}
\centering
\textbf{\small (b) AutoDMP}
\end{minipage}
\begin{minipage}{0.18\textwidth}
\centering
\textbf{\small (c) WireMask-EA}
\end{minipage}
\begin{minipage}{0.18\textwidth}
\centering
\textbf{\small (d) MaskPlace}
\end{minipage}
\begin{minipage}{0.18\textwidth}
\centering
\textbf{\small (e) MaskRegulate}
\end{minipage}
\caption{Placement layouts and congestions of different methods on the eight ICCAD 2015 benchmarks. The congestion results are obtained by \emph{Cadence Innovus}, where red points indicate the congestion critical regions.}
\label{fig:congestions-all}
\end{figure*}

We provide the detail visualizations of placement results of all the methods on all the eight chips from ICCAD 2015. As shown in Figure~\ref{fig:congestions-all}, our proposed MaskRegulate shows significant improvements on congestion metrics. Besides, the placement result of MaskRegulate is much regular than all the other methods. 

\subsection{Additional results.}
We conduct the following additional results to comprehensively show the effectiveness of our MaskRegulate.
1) To verify whether using a better model structure for the RL placer can compare to the regulator, we add comparison with recent proposed ChiPFormer~\cite{lai2023chipformer} under a fair setting. 
2) To further show the generalization ability of our methods, we conduct  generalization experiments on the ISPD 2005 benchmark~\citep{nam2005ispd2005}.
3) To investigate whether MaskRegulate can be used to adjust any initial macro placement solution, we use the pre-trained model to fine-tune other placement results.
These results further demonstrate the competitive results of our proposed MaskRegulate. Detailed discussions are provided in Appendix~\ref{app:exp-chipformer}, \ref{app:ispd}, and \ref{app:fine-tune}, respectively.

\section{Final Remarks}

\textbf{Conclusion.} In this paper, we present a novel RL problem formulation for macro placement, focusing on the development of a macro regulator rather than a placer. Our proposed method, MaskRegulate, demonstrates substantial improvements in chip placement quality by refining existing layouts instead of generating them from scratch. By integrating dense reward signals and emphasizing regularity, our approach effectively addresses the limitations of traditional RL-based placement methods, resulting in superior performance in PPA metrics across various chips. This advancement paves the way for more efficient and effective chip design through RL.

\textbf{Limitations and future work.} This study has several primary limitations: it does not consider the impact of module aspect ratio and area factors on placement; it overlooks global wirelength and timing metrics during the training process; and it does not employ advanced transformer architectures~\citep{lai2023chipformer} to enhance the generalization of the regulator. Chip design inherently involves different preferences, such as the need for compact size in mobile phone chips and larger sizes for computer chips. Therefore, future research should address these challenges and explore efficient methods to obtain a set of chip placements that accommodate different preferences using multi-objective optimization.

\begin{ack}
We thank the reviewers for their insightful and valuable comments. 
This work was supported by the National Science and Technology Major Project (2022ZD0116600), the National Science Foundation of China (62276124), and the Fundamental Research Funds for the Central Universities (14380020).
\end{ack}

\bibliography{main}
\bibliographystyle{abbrv}

\clearpage
\newpage
\appendix

\section{Implementation Details}

\subsection{Benchmarks}~\label{app:benchmarks}

The detailed statistics of our benchmarks are listed in Table~\ref{tab:iccad15}.

\begin{table}[htbp]
\caption{Detailed statistics of the benchmarks.}
\centering
\begin{tabular}{@{}c|ccc@{}}
\toprule
Benchmark & \#Cells   & \#Nets    & \#Pins \\ \midrule
adaptec1  & 210,904   & 221,142     & 944,053    \\
adaptec2  & 254,457   & 266,009   & 1,069,482  \\
adaptec3  & 450,927   & 466,758   & 1,875,039  \\
adaptec4  & 494,716   & 515,951   & 1,912,420  \\
bigblue1  & 277,604   & 284,479   & 1,144,691    \\
bigblue2  & 534,782   & 577,235   & 2,122,282    \\
bigblue3  & 1,095,514 & 1,123,170 & 3,833,218  \\
bigblue4  & 2,169,183   & 2,229,886   & 8,900,078    \\
superblue1 & 1,209,716 & 1215710 & 3,767,494\\ 
superblue3 & 1,213,253 & 1,224,979 & 3,905,321 \\
superblue4 & 795,645 & 802,513 & 2,497,940 \\
superblue5 & 1,086,888 & 1,100,825 & 3,246,878\\
superblue7 & 1,931,639 & 1,933,945 & 6,372,094 \\
superblue10 & 1,876,103 & 1,898,119 & 5,560,506\\
superblue16 & 981,559 & 999,902 & 3,013,268 \\
superblue18 & 768,068 & 771,542 & 2,559,143 \\
\bottomrule
\end{tabular}\label{tab:iccad15}
\end{table}

\subsection{HPWL calculation}~\label{app:wiremask}

\begin{figure*}[htbp]
\centering
\includegraphics[width=0.5\textwidth]{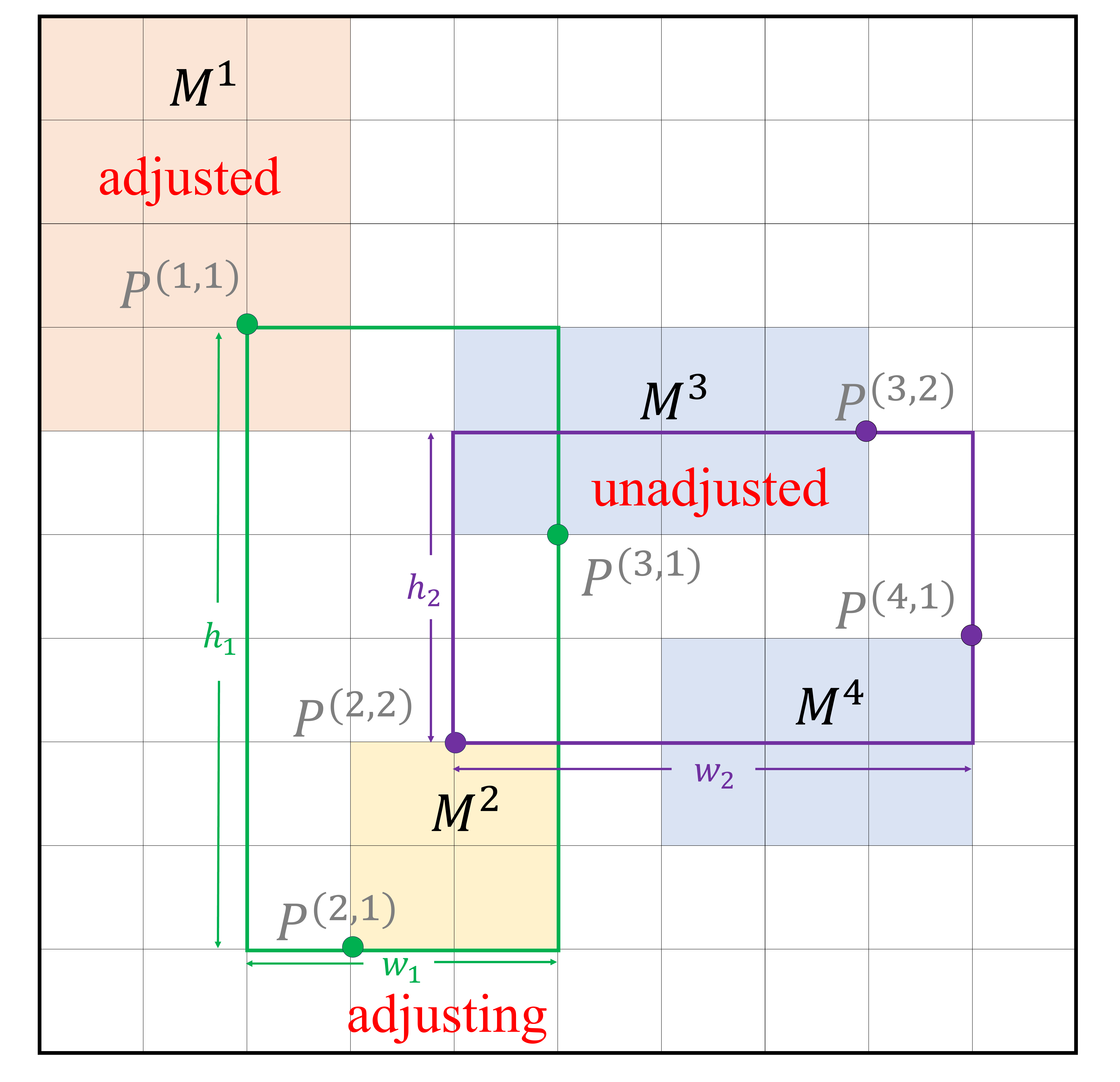} 
\caption{Illustration of chip canvas and calculation of HPWL.}
\label{fig:hpwl-calculation}
\end{figure*}

HPWL (half perimeter wirelength) is an important metric which measures the placement quality before routing. Intuitively, Figure~\ref{fig:hpwl-calculation} illustrates a 2D chip canvas where $M^i$ and $P^{(i,j)}$ denote the $i$-th module to adjust and its $j$-th pin, respectively. Solid boxes in green and purple represent the bounding boxes for two distinct nets on the canvas. In concretely, "Net 1"(in green) connects modules $M^1$, $M^2$ and $M^3$ using wires through pins $P^{(1,1)}$, $P^{(2,1)}$ and $P^{(3,1)}$, while "Net 2"(in purple) connects modules $M^2$, $M^3$ and $M^4$ using wires through pins $P^{(2,2)}$, $P^{(3,2)}$ and $P^{(4,1)}$. As shown in Figure~\ref{fig:illustration-mask}(a), HPWL can be computed as $h_1+w_1+h_2+w_2$.

\subsection{Detailed settings of methods}~\label{app:methods}
We conclude some important settings of different methods. For the four compared methods (i.e., DREAMPlace\footnote{\url{https://github.com/limbo018/DREAMPlace}}, AutoDMP\footnote{\url{https://github.com/NVlabs/AutoDMP}}, WireMask-EA\footnote{\url{https://github.com/lamda-bbo/WireMask-BBO}}, and MaskPlace\footnote{\url{https://github.com/laiyao1/maskplace}}), we use their original implementations.
\begin{itemize}
    \item The size of grids is 224, which is same as the original implementation of MaskPlace~\citep{lai2022maskplace}.
    \item Hyperparameters. 
    \begin{table}[htbp]
        \centering
        \caption{Hyperparameters}
        \begin{tabular}{cccc}
            \toprule
             Configuration & Value & Configuration & Value  \\
             \midrule
             Optimizer & Adam & Learning rate & $2.5\times 10^{-3}$ \\
             Total episode & 1000 & Epoch for update & 10 \\
             Batch size & 64 & Buffer capacity & 5120 \\
             Clip $\epsilon$ & 0.2 & Clip gradient norm & 0.5 \\
             Reward discount $\gamma$ & 0.95 & Mask soft coefficient & 1 \\
             DREAMPlace evaluation number & 3 & Trade-off coefficient $\alpha$ & 0.7 \\
             Grid soft coefficient & 4 & & \\
             \bottomrule
        \end{tabular}
    \end{table}
    \item Neural Network architecture. 
    \begin{table}[htbp]
        \centering
        \caption{Neural Network architecture}
        \begin{tabular}{cccc}
        \toprule
        Block & Layer & Kernel Size & Output Shape  \\ 
        \midrule
        \multirow{3}{*}{Local Mask Fusion} & Conv & $1\times1$ & $(224,224,12)$ \\
         & Conv & $1\times1$ & $(224,224,12)$ \\
         & Conv & $1\times1$ & $(224,224,1)$ \\
        \midrule
        \multirow{2}{*}{Global Mask Encoder} & ResNet-18  & - & $1000$ \\
        & FC & - & 784 \\
        \midrule
        \multirow{5}{*}{Global Mask Decoder} & Deconv & $3\times3$ & $(14,14,8)$ \\
        & Deconv & $3\times3$ & $(28,28,4)$ \\
        & Deconv & $3\times3$ & $(56,56,2)$ \\
        & Deconv & $3\times3$ & $(112,112,1)$ \\
        & Deconv & $3\times3$ & $(224,224,1)$ \\
        \midrule
        Merge & Conv & $1\times 1$ & $(224,224,1)$ \\
        \midrule
        Position Embedding & - & - & 64 \\
        \midrule
        \multirow{3}{*}{FC for value} & FC & - & $64$ \\
        & FC & - & $64$ \\
        & FC & - & $1$ \\
        \bottomrule
        \end{tabular}
    \end{table}
    \item Device.  \\
    CPU: Intel(R) Xeon(R) Gold 6430 \\
    GPU: 4 $\times$ GeForce RTX 4090
\end{itemize}

\section{Additional Experimental Results}~\label{app:additional-results}

\subsection{Ablation studies}~\label{app:exp-ablation}
\textbf{Purely comparison between placer and regulator.} Here, we only change the problem formulation for purely comparing placer and regulator. We implement Vanilla-MaskRegulate, where the only difference to MaskPlace is the problem formulation, and all the other components (e.g., state and reward) are the same. The results in Table~\ref{tab:vanilla-regularmask} clearly demonstrates our motivation, highlighting the advantages of our regulator problem formulation.

\begin{table}[htbp]
\caption{Results of MaskPlace and Vanilla-MaskRegulate on four chips of ICCAD 2015 benchmarks. The only one difference between these two methods is the problem formulation, where all the other components are the same. The best result on each chip is \textbf{bolded}.}
\centering
\begin{tabular}{c|c|c}
\toprule
\multirow{2}{*}{Benchmark}  & MaskPlace & Vanilla-MaskRegulate  \\
          & \multicolumn{2}{c}{Global HPWL (1e8)} \\\midrule
\multirow{1}{*}{superblue1}               & 7.93 $\pm$ 0.06               & \textbf{5.58 $\pm$ 0.05}                          \\
\multirow{1}{*}{superblue3}                     & 9.00 $\pm$ 0.17               & \textbf{8.01 $\pm$ 0.03}                          \\
\multirow{1}{*}{superblue4}                    & 5.28 $\pm$ 0.03               & \textbf{4.31 $\pm$ 0.01}                          \\
\multirow{1}{*}{superblue5}                    & 9.81 $\pm$ 0.03               & \textbf{7.89 $\pm$ 0.02}                         \\\bottomrule
\end{tabular}\label{tab:vanilla-regularmask}
\end{table}

\textbf{MaskRegulate with or without normalization.} The results are shown in Table~\ref{tab:normalization}. Since global HPWL has large scale than regularity, MaskRegulate w/o normalization does not prefer to consider regularity, which is not what we expect. 

\begin{table}[htbp]
\caption{Results of MaskRegulate and MaskRegulate without normalization on the four chips of ICCAD 2015 benchmarks. The best result of each metric on each chip is \textbf{bolded}.}
\centering
\begin{tabular}{c|cc|cc}
\toprule
\multirow{2}{*}{Benchmark}  & \multicolumn{2}{c}{MaskRegulate}   & \multicolumn{2}{|c}{MaskRegulate w/o normalization} \\
                           & Global HPWL (1e8)               & Regularity (1e6)         & Global HPWL (1e8)                        & Regularity (1e6)         \\\midrule
superblue1                 & \textbf{5.77 $\pm$ 0.05} & \textbf{3.31 $\pm$ 0.00} & 5.82 $\pm$ 0.07                   & 3.42 $\pm$ 0.06          \\
superblue3                 & 7.05 $\pm$ 0.03          & \textbf{3.54 $\pm$ 0.00} & \textbf{6.71 $\pm$ 0.03}          & 3.59 $\pm$ 0.01          \\
superblue4                 & 4.15 $\pm$ 0.06          & \textbf{2.18 $\pm$ 0.02} & \textbf{3.99 $\pm$ 0.02}          & 2.49 $\pm$ 0.10          \\
superblue5                 & \textbf{6.94 $\pm$ 0.00} & \textbf{4.23 $\pm$ 0.01} & 7.03 $\pm$ 0.04                   & 4.26 $\pm$ 0.01\\\bottomrule
\end{tabular}~\label{tab:normalization}
\end{table}

\textbf{Training regularity-aware RL placer from scratch.}~\label{app:exp-placer+regularity}
Our proposed RegularMask and regularity-based reward function can also be used to train a RL placer from scratch. We implement MaskPlace+RegularMask and compare it with MaskPlace and MaskRegulate. The results show the advantages of the integration of regularity (between MaskPlace and MaskPlace + RegularMask) and our RL regular formulation (between MaskPlace + RegularMask and MaskRegulate). 

The above ablation results demonstrate the effectiveness of each component of MaskRegulate. 

\begin{table}[htbp]
\caption{Results of MaskPlace, MaskPlace + RegularMask, MaskRegulate on the eight chips of ICCAD 2015 benchmarks. The best result of each metric on each chip is \textbf{bolded}.}
\centering
\resizebox{\columnwidth}{!}{%
\begin{tabular}{c|cc|cc|cc}
\toprule
\multirow{2}{*}{Benchmark} & \multicolumn{2}{c}{MaskPlace} & \multicolumn{2}{|c}{MaskPlace + RegularMask}   & \multicolumn{2}{|c}{MaskRegulate}           \\
                           & Global HPWL (1e8)        & Regularity   (1e6) & Global HPWL (1e8)                & Regularity   (1e6)       & Global HPWL   (1e8)             & Regularity   (1e6)       \\\midrule
superblue1                 & 7.93 $\pm$ 0.06   & 4.40 $\pm$ 0.06    & 7.44 $\pm$ 0.08           & 3.87 $\pm$ 0.06          & \textbf{5.77 $\pm$ 0.05} & \textbf{3.31 $\pm$ 0.00} \\
superblue3                 & 8.90 $\pm$ 0.17   & 4.77 $\pm$ 0.06    & 7.18 $\pm$ 0.05           & \textbf{3.53 $\pm$ 0.02} & \textbf{7.05 $\pm$ 0.03} & 3.54 $\pm$ 0.00          \\
superblue4                 & 5.28 $\pm$ 0.03   & 3.22 $\pm$ 0.03    & 4.49 $\pm$ 0.03           & 2.21 $\pm$ 0.03          & \textbf{4.15 $\pm$ 0.06} & \textbf{2.18 $\pm$ 0.02} \\
superblue5                 & 9.81 $\pm$ 0.03   & 4.86 $\pm$ 0.04    & 7.52 $\pm$ 0.08           & \textbf{4.23 $\pm$ 0.02} & \textbf{6.94 $\pm$ 0.00} & 4.23 $\pm$ 0.01          \\
superblue7                 & 9.99 $\pm$ 0.05   & 4.36 $\pm$ 0.03    & 8.74 $\pm$ 0.09           & 3.07 $\pm$ 0.01          & \textbf{7.90 $\pm$ 0.03} & \textbf{3.03 $\pm$ 0.00} \\
superblue10                & 10.94 $\pm$ 0.21  & 4.75 $\pm$ 0.11    & \textbf{10.90 $\pm$ 0.30} & 4.46 $\pm$ 0.26          & 11.23 $\pm$   0.38       & \textbf{3.90 $\pm$ 0.44} \\
superblue16                & 5.53 $\pm$ 0.04   & 3.08 $\pm$ 0.06    & 5.72 $\pm$ 0.03           & 2.94 $\pm$ 0.06          & \textbf{5.35 $\pm$ 0.05} & \textbf{1.85 $\pm$ 0.02} \\
superblue18                & 3.49 $\pm$ 0.06   & 2.59 $\pm$ 0.08    & 2.89 $\pm$ 0.03           & 1.56 $\pm$ 0.01          & \textbf{2.96 $\pm$ 0.01} & \textbf{1.55 $\pm$ 0.00}\\\bottomrule
\end{tabular}~\label{tab:maskplace+regular}
}
\end{table}

\subsection{Comparison with ChiPFormer}~\label{app:exp-chipformer}
Recently, ChiPFormer~\cite{lai2023chipformer} incorporates an offline learning decision transformer to improve the generalizability. However, we find that even after fine-tuning for the same number of episodes as MaskRegulate, it is still challenging to achieve satisfactory results on ICCAD 2015. Due to the resource-intensive nature of fine-tuning it, we conducted only a partial set of experiments, which is why they were not included in the main paper. In the future, we plan to explore training a generalized Regulator based on the transformer and pre-train it on some chips from ICCAD 2015, and compare it with ChiPFormer that pre-train on the same training chips.

\begin{table}[htbp]
\caption{Results of MaskRegulate and ChiPFormer on eight chips of ICCAD 2015 benchmarks. The best result of each metric on each chip is \textbf{bolded}.}
\centering
\begin{tabular}{c|cc|cc}
\toprule
\multirow{2}{*}{Benchmark}   & \multicolumn{2}{c}{MaskRegulate}            & \multicolumn{2}{|c}{ChiPFormer} \\
\textbf{}   & Global HPWL (1e8)                & Regularity (1e6)         & Global HPWL (1e8)         & Regularity (1e6)   \\\midrule
superblue1  & \textbf{5.77 $\pm$ 0.05}  & \textbf{3.31 $\pm$ 0.00} & 8.09 $\pm$ 0.00    & 4.84 $\pm$ 0.00    \\
superblue3  & \textbf{7.05 $\pm$ 0.03}  & \textbf{3.54 $\pm$ 0.00} & 9.22 $\pm$ 0.02    & 5.13 $\pm$ 0.00    \\
superblue4  & \textbf{4.15 $\pm$ 0.06}  & \textbf{2.18 $\pm$ 0.02} & 5.11 $\pm$ 0.00    & 3.66 $\pm$ 0.00    \\
superblue5  & \textbf{6.94 $\pm$ 0.00}  & \textbf{4.23 $\pm$ 0.01} & 10.97 $\pm$ 0.14   & 5.62 $\pm$ 0.01    \\
superblue7  & \textbf{7.90 $\pm$ 0.03}  & \textbf{3.03 $\pm$ 0.00} & 16.81 $\pm$ 0.09   & 4.78 $\pm$ 0.00    \\
superblue10 & \textbf{11.23 $\pm$ 0.38} & \textbf{3.90 $\pm$ 0.44} & 14.15 $\pm$ 0.17   & 5.18 $\pm$ 0.00    \\
superblue16 & \textbf{5.35 $\pm$ 0.05}  & \textbf{1.85 $\pm$ 0.02} & 5.88 $\pm$ 0.00    & 3.67 $\pm$ 0.00    \\
superblue18 & \textbf{2.96 $\pm$ 0.01}  & \textbf{1.55 $\pm$ 0.00} & 3.57 $\pm$ 0.00    & 3.18 $\pm$ 0.00   \\\bottomrule
\end{tabular}~\label{tab:chipformer}
\end{table}

\subsection{Experiments on ISPD 2005}\label{app:ispd}
We test the generalization on the ISPD 2005 benchmark~\citep{nam2005ispd2005} by directly using the pre-trained models on superblue 1, 3, 4, and 5 (i.e., the same models in Table~\ref{tab:generalization}) of MaskPlace and MaskRegulate to place and regulate the eight chips. As shown in Table~\ref{tab:ISPD2005}, MaskRegulate still outperforms MaskPlace in most cases, demonstrating our superior generalization ability and robustness. 

\begin{table}[htbp]
\caption{Generalization results of proxy metrics on eight chips of ISPD 2005 benchmarks. The best result of each metric on each chip is \textbf{bolded}.}
\centering
\begin{tabular}{c|cc|cc}
\toprule
\multirow{2}{*}{Benchmark}   & \multicolumn{2}{c}{MaskPlace}  & \multicolumn{2}{|c}{MaskRegulate}  \\
\textbf{}   & Global HPWL (1e7)                & Regularity (1e3)         & Global HPWL (1e7)         & Regularity (1e3)  \\
\midrule
adaptec1  & 10.58 $\pm$ 0.07  & 4.68 $\pm$ 0.08  & \textbf{7.75 $\pm$ 0.12}  & \textbf{3.33 $\pm$ 0.00}  \\
adaptec2  & 13.91 $\pm$ 0.22  & 5.35 $\pm$ 0.03  & \textbf{9.53 $\pm$ 0.09}  & \textbf{4.74 $\pm$ 0.00}   \\
adaptec3  & 23.62 $\pm$ 0.25  & 10.04 $\pm$ 0.03 & \textbf{20.40 $\pm$ 0.14}  & \textbf{7.64 $\pm$ 0.01}    \\
adaptec4  & \textbf{23.92 $\pm$ 0.16}  & 11.29 $\pm$ 0.05 & 24.45 $\pm$ 0.06  & \textbf{8.60 $\pm$ 0.02}    \\
bigblue1  & 10.78 $\pm$ 0.01  & 5.20 $\pm$ 0.04 & \textbf{9.34 $\pm$ 0.03}  & \textbf{2.75 $\pm$ 0.01}    \\
bigblue2  & 34.31 $\pm$ 0.31  & \textbf{8.92 $\pm$ 0.06} & \textbf{24.32 $\pm$ 0.46}  & 9.76 $\pm$ 0.01   \\
bigblue3  & 51.53 $\pm$ 0.63  & 10.32 $\pm$ 0.24 & \textbf{36.97 $\pm$ 0.35}  & \textbf{8.08 $\pm$ 0.03}  \\
bigblue4  & 134.97 $\pm$ 2.45  & 17.32 $\pm$ 0.20 & \textbf{93.96 $\pm$ 0.35}  & \textbf{12.38 $\pm$ 0.03} \\ 
\bottomrule
\end{tabular}~\label{tab:ISPD2005}
\end{table}

\subsection{Experiments on fine-tuning existing placement results}\label{app:fine-tune}

To investigate whether MaskRegulate can be used to adjust any initial macro placement solution, we conduct additional experiments to demonstrate this capability. We used the pre-trained model on superblue 1, 3, 4, and 5 (i.e., the same models in Tables~\ref{tab:generalization} and~\ref{tab:ISPD2005}) to adjust different placement results obtained by MaskPlace, AutoDMP, and WireMask-EA. The results are shown in Table~\ref{tab:finetune}. MaskRegulate consistently improves regularity on all four unseen chips and enhances global HPWL on three chips.

\begin{table}[htbp]
\caption{Results of proxy metrics on four chips of ICCAD 2015 benchmarks. We use our policy trained on superblue1, superblue3, superblue4 and superblue5 to finetune the placements gained from MaskPlace, AutoDMP and WireMask-BBO on superblue7, superblue10, superblue16 and superblue18. The left column indicates the Global HPWL (1e8) while the right column indicates the regularity (1e6). The best result of each metric on each chip is \textbf{bolded}.}
\centering
\begin{tabular}{c|cc|cc|cc|cc}
\toprule
Method  & \multicolumn{2}{c}{superblue7}  & \multicolumn{2}{|c}{superblue10} & \multicolumn{2}{|c}{superblue16} & \multicolumn{2}{|c}{superblue18} \\
\midrule
MaskPlace & 9.92 & 4.32 & \textbf{10.55} & 4.87 & 5.46 & 3.02 & 3.40 & 2.57 \\
MaskRegulate + MaskPlace & \textbf{8.53}  & \textbf{3.05} & 11.20  & \textbf{3.25}
& \textbf{5.27}  & \textbf{1.89} & \textbf{2.88}  & \textbf{1.56} \\
\midrule
AutoDMP & 10.68 & 3.96 & 11.57 & 3.73 & 5.87 & 2.56 & 3.02 & 2.29 \\
MaskRegulate + AutoDMP & \textbf{8.28}  & \textbf{3.02} & \textbf{11.39}  & \textbf{3.25} 
& \textbf{5.11}  & \textbf{1.89} & \textbf{2.94}  & \textbf{1.56} \\
\midrule
WireMask-EA & 9.53 & 4.33 & \textbf{10.99} & 4.86 & 5.91 & 3.10 & 3.37 & 2.60 \\
MaskRegulate + WireMask-EA & \textbf{8.54} & \textbf{3.02}  & 12.02  & \textbf{3.25}
 & \textbf{5.20}  & \textbf{1.85} & \textbf{3.01}  & \textbf{1.54} \\
\bottomrule
\end{tabular}~\label{tab:finetune}
\end{table}

\clearpage
\newpage

\section{NeurIPS paper checklist}

\begin{enumerate}

\item {\bf Claims}
    \item[] Question: Do the main claims made in the abstract and introduction accurately reflect the paper's contributions and scope?
    \item[] Answer: \answerYes{}
    \item[] Justification: See the last sentence of abstract and last paragraph of introduction.
    \item[] Guidelines:
    \begin{itemize}
        \item The answer NA means that the abstract and introduction do not include the claims made in the paper.
        \item The abstract and/or introduction should clearly state the claims made, including the contributions made in the paper and important assumptions and limitations. A No or NA answer to this question will not be perceived well by the reviewers. 
        \item The claims made should match theoretical and experimental results, and reflect how much the results can be expected to generalize to other settings. 
        \item It is fine to include aspirational goals as motivation as long as it is clear that these goals are not attained by the paper. 
    \end{itemize}

\item {\bf Limitations}
    \item[] Question: Does the paper discuss the limitations of the work performed by the authors?
    \item[] Answer: \answerYes{} 
    \item[] Justification: See the last paragraph of the paper.
    \item[] Guidelines:
    \begin{itemize}
        \item The answer NA means that the paper has no limitation while the answer No means that the paper has limitations, but those are not discussed in the paper. 
        \item The authors are encouraged to create a separate "Limitations" section in their paper.
        \item The paper should point out any strong assumptions and how robust the results are to violations of these assumptions (e.g., independence assumptions, noiseless settings, model well-specification, asymptotic approximations only holding locally). The authors should reflect on how these assumptions might be violated in practice and what the implications would be.
        \item The authors should reflect on the scope of the claims made, e.g., if the approach was only tested on a few datasets or with a few runs. In general, empirical results often depend on implicit assumptions, which should be articulated.
        \item The authors should reflect on the factors that influence the performance of the approach. For example, a facial recognition algorithm may perform poorly when image resolution is low or images are taken in low lighting. Or a speech-to-text system might not be used reliably to provide closed captions for online lectures because it fails to handle technical jargon.
        \item The authors should discuss the computational efficiency of the proposed algorithms and how they scale with dataset size.
        \item If applicable, the authors should discuss possible limitations of their approach to address problems of privacy and fairness.
        \item While the authors might fear that complete honesty about limitations might be used by reviewers as grounds for rejection, a worse outcome might be that reviewers discover limitations that aren't acknowledged in the paper. The authors should use their best judgment and recognize that individual actions in favor of transparency play an important role in developing norms that preserve the integrity of the community. Reviewers will be specifically instructed to not penalize honesty concerning limitations.
    \end{itemize}

\item {\bf Theory Assumptions and Proofs}
    \item[] Question: For each theoretical result, does the paper provide the full set of assumptions and a complete (and correct) proof?
    \item[] Answer: \answerNA{} 
    \item[] Justification: There is no theoretical results in this paper.
    \item[] Guidelines:
    \begin{itemize}
        \item The answer NA means that the paper does not include theoretical results. 
        \item All the theorems, formulas, and proofs in the paper should be numbered and cross-referenced.
        \item All assumptions should be clearly stated or referenced in the statement of any theorems.
        \item The proofs can either appear in the main paper or the supplemental material, but if they appear in the supplemental material, the authors are encouraged to provide a short proof sketch to provide intuition. 
        \item Inversely, any informal proof provided in the core of the paper should be complemented by formal proofs provided in appendix or supplemental material.
        \item Theorems and Lemmas that the proof relies upon should be properly referenced. 
    \end{itemize}

    \item {\bf Experimental Result Reproducibility}
    \item[] Question: Does the paper fully disclose all the information needed to reproduce the main experimental results of the paper to the extent that it affects the main claims and/or conclusions of the paper (regardless of whether the code and data are provided or not)?
    \item[] Answer: \answerYes{} 
    \item[] Justification: We have provided our code in the supplemental file.
    \item[] Guidelines:
    \begin{itemize}
        \item The answer NA means that the paper does not include experiments.
        \item If the paper includes experiments, a No answer to this question will not be perceived well by the reviewers: Making the paper reproducible is important, regardless of whether the code and data are provided or not.
        \item If the contribution is a dataset and/or model, the authors should describe the steps taken to make their results reproducible or verifiable. 
        \item Depending on the contribution, reproducibility can be accomplished in various ways. For example, if the contribution is a novel architecture, describing the architecture fully might suffice, or if the contribution is a specific model and empirical evaluation, it may be necessary to either make it possible for others to replicate the model with the same dataset, or provide access to the model. In general. releasing code and data is often one good way to accomplish this, but reproducibility can also be provided via detailed instructions for how to replicate the results, access to a hosted model (e.g., in the case of a large language model), releasing of a model checkpoint, or other means that are appropriate to the research performed.
        \item While NeurIPS does not require releasing code, the conference does require all submissions to provide some reasonable avenue for reproducibility, which may depend on the nature of the contribution. For example
        \begin{enumerate}
            \item If the contribution is primarily a new algorithm, the paper should make it clear how to reproduce that algorithm.
            \item If the contribution is primarily a new model architecture, the paper should describe the architecture clearly and fully.
            \item If the contribution is a new model (e.g., a large language model), then there should either be a way to access this model for reproducing the results or a way to reproduce the model (e.g., with an open-source dataset or instructions for how to construct the dataset).
            \item We recognize that reproducibility may be tricky in some cases, in which case authors are welcome to describe the particular way they provide for reproducibility. In the case of closed-source models, it may be that access to the model is limited in some way (e.g., to registered users), but it should be possible for other researchers to have some path to reproducing or verifying the results.
        \end{enumerate}
    \end{itemize}

\item {\bf Open access to data and code}
    \item[] Question: Does the paper provide open access to the data and code, with sufficient instructions to faithfully reproduce the main experimental results, as described in supplemental material?
    \item[] Answer: \answerYes{} 
    \item[] Justification: We have provided our code in the supplemental file.
    \item[] Guidelines:
    \begin{itemize}
        \item The answer NA means that paper does not include experiments requiring code.
        \item Please see the NeurIPS code and data submission guidelines (\url{https://nips.cc/public/guides/CodeSubmissionPolicy}) for more details.
        \item While we encourage the release of code and data, we understand that this might not be possible, so “No” is an acceptable answer. Papers cannot be rejected simply for not including code, unless this is central to the contribution (e.g., for a new open-source benchmark).
        \item The instructions should contain the exact command and environment needed to run to reproduce the results. See the NeurIPS code and data submission guidelines (\url{https://nips.cc/public/guides/CodeSubmissionPolicy}) for more details.
        \item The authors should provide instructions on data access and preparation, including how to access the raw data, preprocessed data, intermediate data, and generated data, etc.
        \item The authors should provide scripts to reproduce all experimental results for the new proposed method and baselines. If only a subset of experiments are reproducible, they should state which ones are omitted from the script and why.
        \item At submission time, to preserve anonymity, the authors should release anonymized versions (if applicable).
        \item Providing as much information as possible in supplemental material (appended to the paper) is recommended, but including URLs to data and code is permitted.
    \end{itemize}

\item {\bf Experimental Setting/Details}
    \item[] Question: Does the paper specify all the training and test details (e.g., data splits, hyperparameters, how they were chosen, type of optimizer, etc.) necessary to understand the results?
    \item[] Answer: \answerYes{} 
    \item[] Justification: We have provided experimental details in Section 4.1 and Appendix A. 
    \item[] Guidelines:
    \begin{itemize}
        \item The answer NA means that the paper does not include experiments.
        \item The experimental setting should be presented in the core of the paper to a level of detail that is necessary to appreciate the results and make sense of them.
        \item The full details can be provided either with the code, in appendix, or as supplemental material.
    \end{itemize}

\item {\bf Experiment Statistical Significance}
    \item[] Question: Does the paper report error bars suitably and correctly defined or other appropriate information about the statistical significance of the experiments?
    \item[] Answer: \answerYes{} 
    \item[] Justification: We have reported the error bars in our experiments. Please see Table 1.
    \item[] Guidelines:
    \begin{itemize}
        \item The answer NA means that the paper does not include experiments.
        \item The authors should answer "Yes" if the results are accompanied by error bars, confidence intervals, or statistical significance tests, at least for the experiments that support the main claims of the paper.
        \item The factors of variability that the error bars are capturing should be clearly stated (for example, train/test split, initialization, random drawing of some parameter, or overall run with given experimental conditions).
        \item The method for calculating the error bars should be explained (closed form formula, call to a library function, bootstrap, etc.)
        \item The assumptions made should be given (e.g., Normally distributed errors).
        \item It should be clear whether the error bar is the standard deviation or the standard error of the mean.
        \item It is OK to report 1-sigma error bars, but one should state it. The authors should preferably report a 2-sigma error bar than state that they have a 96\% CI, if the hypothesis of Normality of errors is not verified.
        \item For asymmetric distributions, the authors should be careful not to show in tables or figures symmetric error bars that would yield results that are out of range (e.g. negative error rates).
        \item If error bars are reported in tables or plots, The authors should explain in the text how they were calculated and reference the corresponding figures or tables in the text.
    \end{itemize}

\item {\bf Experiments Compute Resources}
    \item[] Question: For each experiment, does the paper provide sufficient information on the computer resources (type of compute workers, memory, time of execution) needed to reproduce the experiments?
    \item[] Answer: \answerYes{} 
    \item[] Justification: This information is provided in Appendix A.
    \item[] Guidelines:
    \begin{itemize}
        \item The answer NA means that the paper does not include experiments.
        \item The paper should indicate the type of compute workers CPU or GPU, internal cluster, or cloud provider, including relevant memory and storage.
        \item The paper should provide the amount of compute required for each of the individual experimental runs as well as estimate the total compute. 
        \item The paper should disclose whether the full research project required more compute than the experiments reported in the paper (e.g., preliminary or failed experiments that didn't make it into the paper). 
    \end{itemize}
    
\item {\bf Code Of Ethics}
    \item[] Question: Does the research conducted in the paper conform, in every respect, with the NeurIPS Code of Ethics \url{https://neurips.cc/public/EthicsGuidelines}?
    \item[] Answer: \answerYes{} 
    \item[] Justification: We have read the NeurIPS Code of Ethics and follow it.
    \item[] Guidelines:
    \begin{itemize}
        \item The answer NA means that the authors have not reviewed the NeurIPS Code of Ethics.
        \item If the authors answer No, they should explain the special circumstances that require a deviation from the Code of Ethics.
        \item The authors should make sure to preserve anonymity (e.g., if there is a special consideration due to laws or regulations in their jurisdiction).
    \end{itemize}

\item {\bf Broader Impacts}
    \item[] Question: Does the paper discuss both potential positive societal impacts and negative societal impacts of the work performed?
    \item[] Answer: \answerYes{} 
    \item[] Justification: The chip is the core productivity of modern society. Our method proposes a more efficient way of using reinforcement learning for macro placement of chips, which has the potential to enhance the quality of chip design.
    \item[] Guidelines:
    \begin{itemize}
        \item The answer NA means that there is no societal impact of the work performed.
        \item If the authors answer NA or No, they should explain why their work has no societal impact or why the paper does not address societal impact.
        \item Examples of negative societal impacts include potential malicious or unintended uses (e.g., disinformation, generating fake profiles, surveillance), fairness considerations (e.g., deployment of technologies that could make decisions that unfairly impact specific groups), privacy considerations, and security considerations.
        \item The conference expects that many papers will be foundational research and not tied to particular applications, let alone deployments. However, if there is a direct path to any negative applications, the authors should point it out. For example, it is legitimate to point out that an improvement in the quality of generative models could be used to generate deepfakes for disinformation. On the other hand, it is not needed to point out that a generic algorithm for optimizing neural networks could enable people to train models that generate Deepfakes faster.
        \item The authors should consider possible harms that could arise when the technology is being used as intended and functioning correctly, harms that could arise when the technology is being used as intended but gives incorrect results, and harms following from (intentional or unintentional) misuse of the technology.
        \item If there are negative societal impacts, the authors could also discuss possible mitigation strategies (e.g., gated release of models, providing defenses in addition to attacks, mechanisms for monitoring misuse, mechanisms to monitor how a system learns from feedback over time, improving the efficiency and accessibility of ML).
    \end{itemize}
    
\item {\bf Safeguards}
    \item[] Question: Does the paper describe safeguards that have been put in place for responsible release of data or models that have a high risk for misuse (e.g., pretrained language models, image generators, or scraped datasets)?
    \item[] Answer: \answerNA{} 
    \item[] Justification: Our paper has no such risks.
    \item[] Guidelines:
    \begin{itemize}
        \item The answer NA means that the paper poses no such risks.
        \item Released models that have a high risk for misuse or dual-use should be released with necessary safeguards to allow for controlled use of the model, for example by requiring that users adhere to usage guidelines or restrictions to access the model or implementing safety filters. 
        \item Datasets that have been scraped from the Internet could pose safety risks. The authors should describe how they avoided releasing unsafe images.
        \item We recognize that providing effective safeguards is challenging, and many papers do not require this, but we encourage authors to take this into account and make a best faith effort.
    \end{itemize}

\item {\bf Licenses for existing assets}
    \item[] Question: Are the creators or original owners of assets (e.g., code, data, models), used in the paper, properly credited and are the license and terms of use explicitly mentioned and properly respected?
    \item[] Answer: \answerYes{}{} 
    \item[] Justification: We cite the original paper and used dataset.
    \item[] Guidelines:
    \begin{itemize}
        \item The answer NA means that the paper does not use existing assets.
        \item The authors should cite the original paper that produced the code package or dataset.
        \item The authors should state which version of the asset is used and, if possible, include a URL.
        \item The name of the license (e.g., CC-BY 4.0) should be included for each asset.
        \item For scraped data from a particular source (e.g., website), the copyright and terms of service of that source should be provided.
        \item If assets are released, the license, copyright information, and terms of use in the package should be provided. For popular datasets, \url{paperswithcode.com/datasets} has curated licenses for some datasets. Their licensing guide can help determine the license of a dataset.
        \item For existing datasets that are re-packaged, both the original license and the license of the derived asset (if it has changed) should be provided.
        \item If this information is not available online, the authors are encouraged to reach out to the asset's creators.
    \end{itemize}

\item {\bf New Assets}
    \item[] Question: Are new assets introduced in the paper well documented and is the documentation provided alongside the assets?
    \item[] Answer: \answerYes{} 
    \item[] Justification: We have provided our code and models in our supplymental file.
    \item[] Guidelines:
    \begin{itemize}
        \item The answer NA means that the paper does not release new assets.
        \item Researchers should communicate the details of the dataset/code/model as part of their submissions via structured templates. This includes details about training, license, limitations, etc. 
        \item The paper should discuss whether and how consent was obtained from people whose asset is used.
        \item At submission time, remember to anonymize your assets (if applicable). You can either create an anonymized URL or include an anonymized zip file.
    \end{itemize}

\item {\bf Crowdsourcing and Research with Human Subjects}
    \item[] Question: For crowdsourcing experiments and research with human subjects, does the paper include the full text of instructions given to participants and screenshots, if applicable, as well as details about compensation (if any)? 
    \item[] Answer: \answerNA{} 
    \item[] Justification: This work does not involve crowdsourcing nor research with human subjects.
    \item[] Guidelines:
    \begin{itemize}
        \item The answer NA means that the paper does not involve crowdsourcing nor research with human subjects.
        \item Including this information in the supplemental material is fine, but if the main contribution of the paper involves human subjects, then as much detail as possible should be included in the main paper. 
        \item According to the NeurIPS Code of Ethics, workers involved in data collection, curation, or other labor should be paid at least the minimum wage in the country of the data collector. 
    \end{itemize}

\item {\bf Institutional Review Board (IRB) Approvals or Equivalent for Research with Human Subjects}
    \item[] Question: Does the paper describe potential risks incurred by study participants, whether such risks were disclosed to the subjects, and whether Institutional Review Board (IRB) approvals (or an equivalent approval/review based on the requirements of your country or institution) were obtained?
    \item[] Answer: \answerNA{} 
    \item[] Justification: This work does not involve crowdsourcing nor research with human subjects.
    \item[] Guidelines:
    \begin{itemize}
        \item The answer NA means that the paper does not involve crowdsourcing nor research with human subjects.
        \item Depending on the country in which research is conducted, IRB approval (or equivalent) may be required for any human subjects research. If you obtained IRB approval, you should clearly state this in the paper. 
        \item We recognize that the procedures for this may vary significantly between institutions and locations, and we expect authors to adhere to the NeurIPS Code of Ethics and the guidelines for their institution. 
        \item For initial submissions, do not include any information that would break anonymity (if applicable), such as the institution conducting the review.
    \end{itemize}

\end{enumerate}

\end{document}